\documentclass[10pt,journal,onecolumn,compsoc]{IEEEtran}
\def\eg{\emph{e.g.\ }} 
\def\ie{\emph{i.e.\ }} 
 
\def\etc{\emph{etc.\ }} 
 
\def\etal{\emph{et al.\ }}

\ifCLASSOPTIONcompsoc
  \usepackage[nocompress]{cite}
\else
  \usepackage{cite}
\fi
\usepackage{subfigure}
\ifCLASSINFOpdf
   \usepackage[pdftex]{graphicx}
\else
\fi
\usepackage{amsmath}
\usepackage{amssymb}
\usepackage{algorithm}  
\usepackage{algorithmic}

\usepackage{array}
\usepackage{booktabs}
\newcommand{\tabincell}[2]{\begin{tabular}{@{}#1@{}}#2\end{tabular}}  
\begin{document}
%
\title{Semi-tensor Product-based Tensor Decomposition for Neural Network Compression}
%
%
%
%

\author{Hengling Zhao, Yipeng~Liu,~\IEEEmembership{Senior Member,~IEEE}, Xiaolin ~Huang,~\IEEEmembership{Senior Member,~IEEE}  Ce~Zhu,~\IEEEmembership{Fellow,~IEEE}, 
\IEEEcompsocitemizethanks{\IEEEcompsocthanksitem This work was supported by the National Natural Science Foundation of China (NSFC) under Grant 61977046, Grant 6220106011 and Grant U19A2052. 
\IEEEcompsocthanksitem H. Zhao, Y. Liu and C. Zhu are with the School of Information and Communication
Engineering, University of Electronic Science and Technology of China (UESTC), Chengdu 611731, China. e-mail: 
yipengliu@uestc.edu.cn.
\IEEEcompsocthanksitem X. Huang is with the Department of Automation, Shanghai Jiao Tong University, Shanghai 200240, China. e-mail: xiaolinhuang@sjtu.edu.cn}}

\IEEEtitleabstractindextext{%
\begin{abstract}
The existing tensor networks adopt conventional matrix product for connection. The classical matrix product requires strict dimensionality consistency between factors, which can result in redundancy in data representation. In this paper, the semi-tensor product is used to generalize classical matrix product based mode product to semi-tensor mode product. As it permits the connection of two factors with different dimensionality, more flexible and compact tensor decompositions can be obtained with smaller sizes of factors. Tucker decomposition, Tensor Train (TT) and Tensor Ring (TR) are common decompositions for low rank compression of deep neural networks. The semi-tensor product is applied to these tensor decompositions to obtained their generalized versions, i.e., semi-tensor Tucker decomposition (STTu), semi-tensor train (STT) and semi-tensor ring (STR). Experimental results show the STTu, STT and STR achieve higher compression factors than the conventional tensor decompositions with the same accuracy but less training times in ResNet and WideResNet compression. With 2\% accuracy degradation, the TT-RN (rank = 14) and the TR-WRN (rank = 16) only obtain 3 times and 99t times compression factors while the STT-RN (rank = 14) and the STR-WRN (rank = 16) achieve 9 times and 179 times compression factors, respectively.

\end{abstract}

}

\maketitle

\IEEEdisplaynontitleabstractindextext

%
\IEEEpeerreviewmaketitle

\IEEEraisesectionheading{\section{Introduction}\label{sec:introduction}}

%
%
%
%
\IEEEPARstart{D}{eep} neural networks (DNNs) have achieved  great performance on many tasks, such as computer vision \cite{long2015fully, kim2014convolutional, tran2015learning}, natural language processing \cite{goldberg2016primer, tai2015improved}, \etc Its success comes from massive data and complicated structure. That means we need to train a huge number of parameters, which is expensive in the view of storage, computation, and time. 
Therefore, compression of DNNs becomes to a key issue for application on 
resource-constrained devices, such as mobile phones and wearable devices \cite{kim2015compression, lane2015early}. 
The current  neural networks compression can be realized mainly by pruning \cite{li2016pruning, liu2017learning}, quantization \cite{jacob2018quantization, rastegari2016xnor}, knowledge distillation \cite{hinton2015distilling}, designing lightweight neural networks  \cite{howard2017mobilenets, ma2018shufflenet, Huang_2018_CVPR}, and low-rank approximation \cite{jaderberg2014speeding, novikov2015tensorizing}. These methods could be investigated parallelly and used simultaneously. In this paper, we focus on low-rank approximation way.

Low-rank approximation 
utilizes tensor decomposition to get a compact representation of the original weight tensors. Generally, it 
can be divided into two categories. One decomposes pre-trained models to get initial low-rank approximations and retrains them to recover the performance of the original network \cite{guo2018network, jaderberg2014speeding, kim2015compression, phan2020stable, zhang2015accelerating, kossaifi2019t}, and the other directly trains a new network whose structure is in low rank tensor form \cite{cao2017tensor, novikov2015tensorizing, wang2018wide}. Compared with the former group (\ie, pre-training and fine-tuning), the latter one ( named as tensor neural networks, TNNs) is an end-to-end processing, which is easier to train. 
Inspired by the tensor decomposition,  the architecture of DNNs could be changed, 
resembling some lightweight convolutional neural networks.
For example, Tucker decomposition uses pointwise convolutions to reduce the number of input channels of the 2D convolution and raises the number of channels after the 2D convolution.

Different tensor decompositions result in different TNNs,
\eg, CP-Nets \cite{lebedev2014speeding}, Tucker-Nets \cite{kossaifi2020tensor, chien2017tensor}, HT-Nets \cite{yin2020compressing}, BT-Nets \cite{ye2020block}, TT-Nets \cite{novikov2015tensorizing, garipov2016ultimate} and TR-Nets \cite{wang2018wide} corresponding to CANDECOMP/PARAFAC  (CP) decomposition \cite{hitchcock1927expression}, Tucker decomposition \cite{tucker1963implications}, Hierarchical Tucker (HT) decomposition \cite{hackbusch2009new}, Block-Term (BT) decomposition \cite{doi:10.1137/070690729}, Tensor Train (TT) \cite{oseledets2011tensor} and Tensor Ring (TR) \cite{zhao2016tensor}, respectively.  
Tjandra \etal \cite{tjandra2018tensor} show that TT-format has a better performance compared with Tucker-format with the same number of parameters required in the recurrent neural network (RNN). Experimental results in \cite{ye2018learning} and \cite{yin2020compressing} 
verify that the performance of RNN compressed by BT decomposition and HT decomposition is better than that of TT decomposition. The above conclusions are all based on a fully-connected layer. But 
the current dominating structures in DNNs are convolutional layers \cite{He_2016_CVPR, simonyan2015deep, Szegedy_2015_CVPR, howard2017mobilenets}. Wu \etal \cite{WU2020309} theoretically and experimentally discover that the HT-format has better performance on compressing weight matrices, while the TT-format is more suitable for compressing convolutional kernels. Besides, Wang \etal \cite{wang2018wide} propose TR-Nets which can significantly compress both the fully connected layers and the convolutional layers.

The construction of above-mentioned tensor decompositions are all based on the mode-$n$ products or tensor contractions. These tensor products are natural generalizations of conventional matrix product but are not very efficient in compression. Conventional matrix product and its tensor counterparts restrict the dimensions between two multipliers, which considerably reduces the flexibility of tensor decomposition and its applicability to specific problems. In this paper, we try to explore more compact low-rank tensor formats. The main idea is to build novel tensor decomposition forms by replacing conventional matrix product by semi-tensor product (STP, \cite{cheng2007survey}),
resulting a general method for various tensor decompositions, named STP-tensor decomposition. STP is a generalization of the conventional matrix product, which 
relaxes the dimension matching condition and hence largely enhances flexibility. Specifically, extending to tensor version, the semi-tensor mode-$n$ product and semi-tensor contraction can be obtained. And the connection way between the factor tensors in tensor networks can be accordingly updated, so that the flexibility of tensor networks is enhanced.

In fact, most existing tensor decompositions can be extended into STP versions. In this paper, we choose Tucker decomposition, TT decomposition and TR decomposition as our baseline. Tucker decomposition is a classical decomposition form, but its number of parameters scale exponentially with the tensor orders. TT and TR decomposition alleviate the problem of dimensionality curse. TR decomposition enjoys enhanced expression and compression capability. It can be viewed as a generalization of CP decomposition and TT decomposition. Wang \etal \cite{wang2017efficient} demonstrate that this generalized factorization is extremely expressive, especially in preserving spatial features, and they propose TR-Nets to significantly compress both the fully-connected layers and the convolutional layers. Novikov \etal \cite{novikov2015tensorizing} define the TT-representation of matrices where TT-factors are forth-order tensors, where the input dimension and the output dimension exist in the same factor. Thus, the number of TT-factors is half of that of TR-factors for factorizing the same tensor.

Driven by this, the semi-tensor ring (STR) decomposition is developed by replacing the connection way within the tensor ring network with the STP. Benefiting from the flexibility of STP, the dimensionality of involved latent factors does not have to match each other. It greatly improves the flexibility of tensor ring network, reduces the number of parameters in need, and accelerate the subsequent data processing. We also define the semi-tensor Tucker (STTu) decomposition and semi-tensor train (STT) decomposition which are the generalization of Tucker decomposition and TT decomposition respectively.

The newly defined tensor networks are used to deep neural network compression. Specifically, we employ the resulting compact tensor networks (\ie STTu, STT and STR) for TNNs compression, namely STP-Nets. We mainly consider two cases, \ie the compression of the fully connected layers and the convolutional layers. The required number of model parameters and computational complexity are compared theoretically and experimentally with existing popular TNNs (\ie Tucker-Nets, TT-Nets, TR-Nets). Furthermore, the effectiveness of proposed STP-Nets is verified on both simple neural network LeNet-5 and deeper neural networks ResNet (RN) and WideResNet (WRN). Experimental results show that the STTu, STT and STR all achieve higher compression factor (CF) than the corresponding conventional tensor decompositions in the same accuracy with less training time in ResNet and WideResNet compression. With 2\% accuracy degradation, the TT-RN ($R = 14$) and the TR-WRN ($R = 16$) only obtain $3 \times$ and $99 \times $ CFs while the STT-RN ($R = 14$) and the STR-WRN ($R = 16$) achieve $9 \times$ and $179 \times$ CFs, respectively. 

In the rest of this paper, section \ref{sec:2} describes tensor diagram notations and the basis about tensor neural networks and semi-tensor product. We firstly introduce STP-tensor decomposition in Section \ref{sec:3}. Then we apply STP-tensor decomposition to neural network compression in Section \ref{sec:4}. Finally,the experimental results show that the newly obtained TNNs outperform the state-of-the-art ones in compressibility without significant performance degradation in Section \ref{sec:5}.

 


\section{Notations and Preliminaries}
\label{sec:2}
\subsection{Tensor Notation}

\begin{figure}[ht]
\vskip 0.1in
\begin{center}
\subfigure[matrix product diagram]{
\label{fig:TDN1} 
\includegraphics[width=0.3\columnwidth]{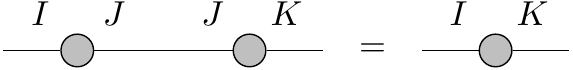}}
\hspace{0.1in}
\subfigure[tensor contraction diagram]{
\label{fig:TDN2} 
\includegraphics[width=0.3\columnwidth]{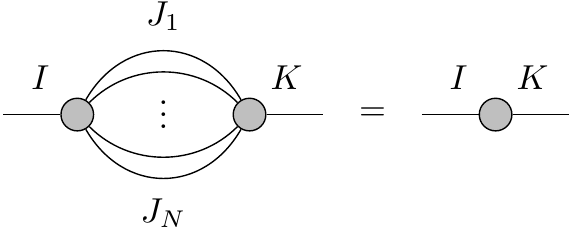}}
\subfigure[tensor self-contraction diagram]{
\label{fig:TDN3} 
\includegraphics[width=0.3\columnwidth]{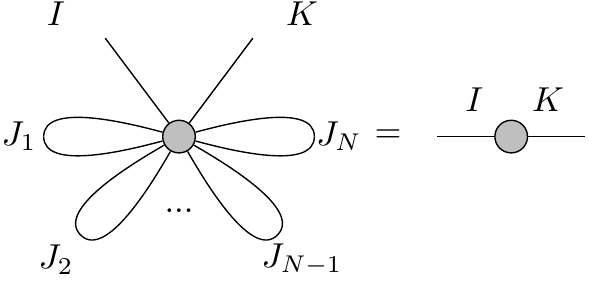}}
\caption{Tensor diagrams. }
\label{fig:TDN}
\end{center}
\vskip -0.2in
\end{figure}

In this paper, we  follow the tensor diagram notation given by \cite{orus2014practical} and shown in 
Fig. \ref{fig:TDN}, 
where node and edge weights represent tensor objects and their dimensionalities, respectively. The connection between two matrices indicates the matrix product of them; the connection between two tensors indicates the tensor contraction of them and the lines correspond to contracted indices.

The matrix product of $\mathbf{X} \in \mathbb{R}^{I\times J}$ and $\mathbf{Y} \in \mathbb{R}^{J\times K}$ is expressed as 
\begin{equation}
\mathbf{Z}=\mathbf{X} \mathbf{Y},
\end{equation}
and the element-wise expression is
\begin{equation}
\mathbf{Z}(i,k) = \Sigma_{j=1}^J \mathbf{X}(i,j) \mathbf{Y}(j,k),
\end{equation}
where $\mathbf{X}(i,j)$ is the element in the $i$-th row and the $j$-th column of $\mathbf{X}$, $\mathbf{Y}(j,k)$  is the element in the $j$-th row and the $k$-th column of $\mathbf{Y}$ and $\mathbf{Z}(i,k)$  is the element in the $i$-th row and the $k$-th column of $\mathbf{Z} \in \mathbb{R}^{I\times K}$. Its tensor diagram illustration is shown in Fig. \ref{fig:TDN1}.

The tensor contraction of $\mathcal{X} \in \mathbb{R}^{I \times J_1 \times \cdots \times J_N}$ and $\mathcal{Y} \in \mathbb{R}^{J_1 \times \cdots \times J_N \times K}$, along the same indices \{$J_1,J_2,\cdots,J_N$\}, is expressed as 
\begin{equation}
\mathbf{Z}= \langle {\mathcal{X}, \mathcal{Y}}\rangle_N \in \mathbb{R}^{I \times K},
\end{equation}
 and the element-wise expression is
\begin{equation}
\mathbf{Z}(i,k) = \Sigma_{j_1,\cdots,j_N}^{J_1,\cdots,J_N} \mathcal{X}(i,j_1, \cdots, j_N) \mathcal{Y}(j_1, \cdots, j_N,k).
\end{equation}
The tensor self-contraction of $\mathcal{X} \in \mathbb{R}^{I \times J_1 \times J_1 \times \cdots \times J_N \times J_N \times K}$, along the same indices \{$J_1,J_2,\cdots,J_N$\}, is expressed as 
\begin{equation}
\mathbf{Z}= \langle {\mathcal{X}}\rangle_N \in \mathbb{R}^{I \times K},
\end{equation}
and the element-wise expression is
\begin{equation}
\mathbf{Z}(i,k) = \Sigma_{j_1,\cdots,j_N}^{J_1,\cdots,J_N} \mathcal{X}(i,j_1,j_1,\cdots,j_N, j_N,k).
\end{equation}
Their tensor diagram notations are shown in Fig. \ref{fig:TDN2} and  Fig. \ref{fig:TDN3}, respectively.

\subsection{Tensor Neural Networks}

In fully connected layers, the linear transformation is formulated as 
\begin{equation}
\mathbf{y} = \mathbf{W}\mathbf{x}+\mathbf{b},
\label{equ:FCL}
\end{equation}
where $\mathbf{x}\in \mathbb{R}^{I}$, $\mathbf{y}\in \mathbb{R}^{O}$ and $\mathbf{W} \in \mathbb{R}^{O\times I}$  are the input, output and weight in matrices of the layer, respectively.

\begin{figure}[ht]
\vskip 0.1in
\begin{center}
\includegraphics[width=0.5\columnwidth]{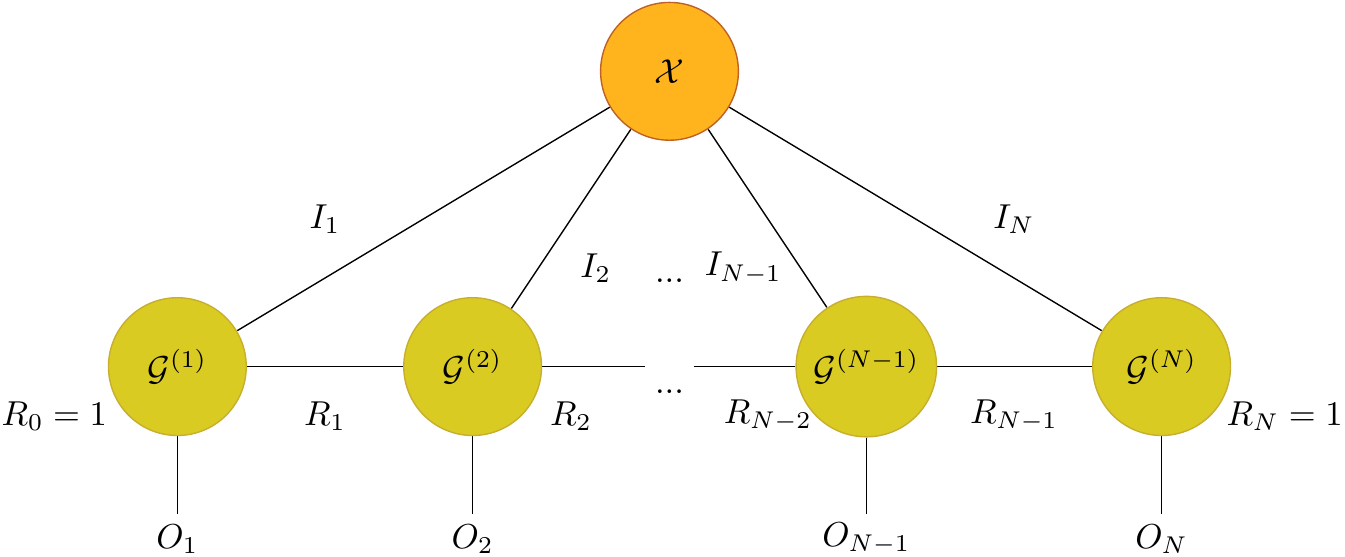} 
\caption{Tensor diagram of TT layers.}
\label{fig:TTN}
\end{center}
\vskip -0.2in
\end{figure}

In tensor neural networks, the input, output and weight are tensors as $\mathcal{X}\in \mathbb{R}^{I_1\times I_2\times \cdots \times I_N}$, $\mathcal{Y}\in \mathbb{R}^{O_1\times O_2\times \cdots \times O_{\hat{N}}}$ and $\mathcal{W} \in \mathbb{R}^{O_1\times \cdots \times O_{\hat{N}}\times I_1\times \cdots \times I_N}$. The aforementioned layers are replaced with 
\begin{equation}
\mathcal{Y} = \langle{\mathcal{W}, \mathcal{X}}\rangle_N+\mathcal{B},
\label{equ:TCL}
\end{equation}
namely tensor regression layers. Considering the parameters redundancy and sharing, the coefficient tensor $\mathcal{W}$ often admits low rank property. For example, when $N=\hat{N}$, Novikov \etal  \cite{novikov2015tensorizing} approximate $\mathcal{W}$ by $N$ TT-cores ($\{\mathcal{G}^{(n)}\}_{n=1}^{N} \in \mathbb{R}^{R_{n-1}\times I_n\times O_n\times R_{n}},R_0=R_{N}=1$ ) as follows
\begin{equation}
\begin{split}
&\mathcal{W}(o_1,\cdots,o_N,i_1,\cdots,i_N) \\= &\mathcal{G}^{(1)}(:,i_1,o_1,:)\cdots \mathcal{G}^{(N)}(:,i_N,o_N,:).
\end{split}
\end{equation}
The tensor regression layers are named TT layers. Fig. \ref{fig:TTN} gives the structure of TT layers using tensor diagram. In addition, some other factorization forms are also exhibited to be effective to compress fully connected layers or convolutional layers, such as Tucker decomposition and tensor ring (TR). The resulting networks include the tensor contraction layer (TRL) \cite{chien2017tensor}, tensor factorized layer (TFL) using the factor matrices \cite{kossaifi2020tensor}, and the TR-Nets \cite{wang2017efficient}.

\subsection{Semi-Tensor Product}

STP is a generalization of the conventional matrix product, which imposes no requirement on the dimensionality of involved matrices or tensors. In other words, the two matrices involved can be in any of the following cases.
\begin{enumerate}
\item The involved two matrices satisfy dimension matching condition which is the same as conventional matrix product.
\item The number of columns in the first matrix is proportional to the number of rows in the second matrix.
\item They are in arbitrary sizes. 
\end{enumerate}

\begin{figure}[ht]
\vskip 0.1in
\begin{center}
\includegraphics[width=0.35\columnwidth]{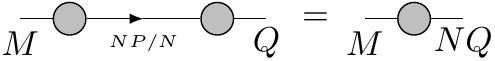}
\caption{Tensor diagram of STP. $NP$ is the number of columns in the first matrix and $N$ represents the ratio between connected dimensionality, called semi-tensor product ratio.}
\label{fig:STP}
\end{center}
\vskip -0.2in
\end{figure}

In this paper, we will focus on the most popular one, \ie case 2). Letting $\mathbf{x}\in\mathbb{R}^{1\times NP}$, $\mathbf{w}\in\mathbb{R}^{P}$, we can split $\mathbf{x}$ into $P$ equal-sized blocks as $\mathbf{x}^1, \mathbf{x}^2,\cdots, \mathbf{x}^P \in \mathbb{R}^{1\times N}$. Then left semi-tensor product (LSTP) denoted by $\ltimes$ performs as follows:
\begin{equation}
\left\{
             \begin{array}{lr}
             \mathbf{x}\ltimes \mathbf{w}= \Sigma_{p=1}^{P}\mathbf{x}^p \mathbf{w}(p) \in \mathbb{R}^{1\times N}, &  \\
             \mathbf{w}^\mathrm{T}\ltimes \mathbf{x}^\mathrm{T} = \Sigma_{p=1}^{P}\mathbf{w}(p) (\mathbf{x}^p)^\mathrm{T} \in \mathbb{R}^N. &  
             \end{array}
\right.
\label{eq:LSTP}
\end{equation}
Similarly, STP of two matrices $\mathbf{X} \in \mathbb{R}^{M\times NP}$  and $\mathbf{W} \in \mathbb{R}^{P\times Q}$ is formulated as 
\begin{equation}
\mathbf{Y} = \mathbf{X} \ltimes \mathbf{W},
\end{equation}
where $\mathbf{Y}$ consists of $M\times Q$ blocks $\mathbf{Y}^{mq} \in \mathbb{R}^{1\times N}$ and each block is
\begin{equation}
\mathbf{Y}^{mq} = \mathbf{X}(m,:) \ltimes \mathbf{W}(:,q),
\end{equation}
where $\mathbf{X}(m,:) $ is the $m$-th row of $\mathbf{X}$ and  $\mathbf{W}(:,q)$ is the $q$-th column of $\mathbf{W}$, $m=1, 2, \cdots, M, \quad q=1, 2, \cdots, Q$. Its tensor diagram notation is shown in Fig. \ref{fig:STP}, the notation `$\blacktriangleright$' on the connection line represents STP and the dimension of the tensor on the left of the arrow along the connecting line is $N$ times that of the tensor on the right.

Because of its flexibility, STP has been applied in many fields. In compressed sensing technique, Fu and Li \cite{fu2018semi} replace the conventional matrix product with STP (in case 2)) in sampling model. The second dimension of sensing matrix $\mathbf{A}$ is factor with the dimension of signal $\mathbf{x}$. For visual question answering, Bai \etal  \cite{bai2020bilinear} apply STP operation (in case 2)) to multi-modal information fusion.  STP is a block-wise operation without dimension constraint.

\subsection{Parameters Sharing}

\begin{figure}[ht]
\vskip 0.1in
\begin{center}
\subfigure[A dense layer]{
\label{fig:dense} 
\includegraphics[width=0.25\columnwidth]{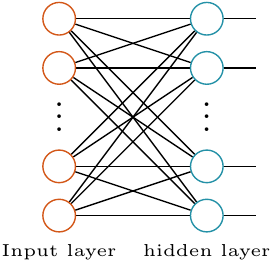}}
\hspace{0.1in}
\subfigure[An STP-layer]{
\label{fig:stp} 
\includegraphics[width=0.25\columnwidth]{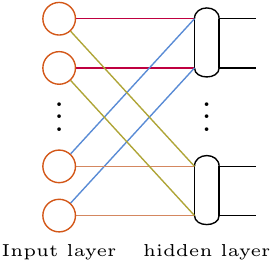}}
\caption{Diagram of two kind of fully-connected layers}
\label{fig:FCL2}
\end{center}
\vskip -0.2in
\end{figure}

The essence of STP in case 2 is the parameters sharing. We use the dense layer in the neural network to illustrate the benefit of this property on the compression parameters. In Eq. \ref{eq:LSTP}, when $\mathbf{x}$ represents an input data and $\mathbf{w}$ represents the corresponding weight of one neuron, STP lets $n$ elements in the input vector share the same weight value. The STP layer can be formulated as $\mathbf{Y} = \mathbf{X} \ltimes \mathbf{W}+\mathbf{b}$, where $\mathbf{X} \in \mathbb{R}^{m\times np}$ is the input, $\mathbf{W} \in \mathbb{R}^{p\times q}$ is the weight matrix, $\mathbf{b} \in \mathbb{R}^{q}$, and $\mathbf{Y} \in \mathbb{R}^{m\times nq}$ is the output.

\begin{figure*}[ht]
\begin{center}
\includegraphics[width=0.9\columnwidth]{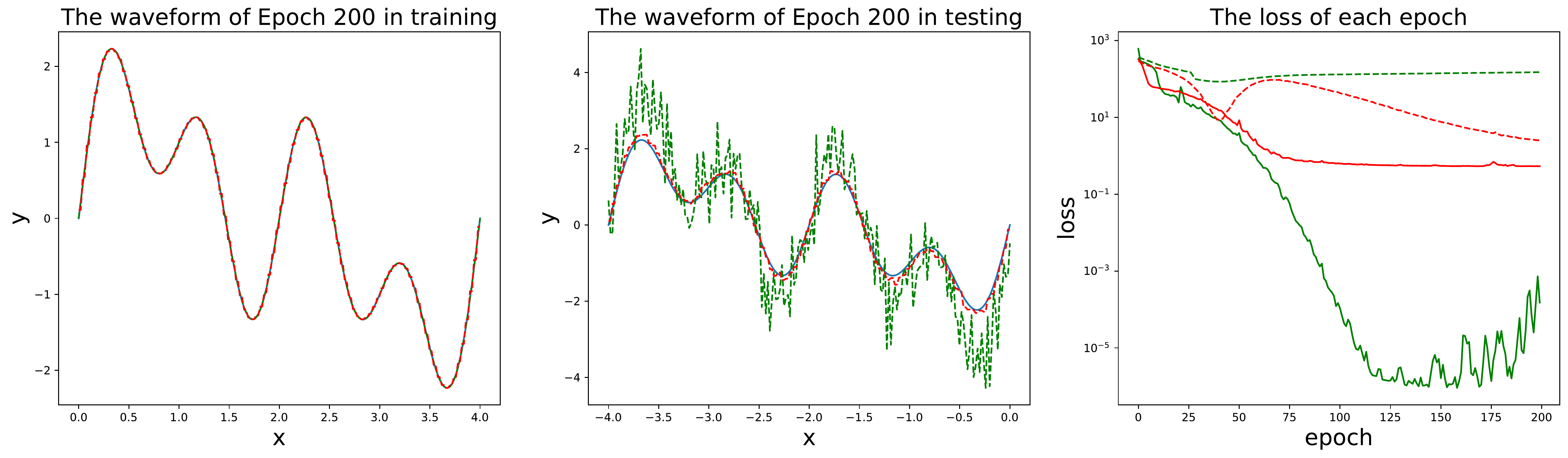} 
\caption{The results of using the base-Net and the STP-Net to approximate a periodic function. The red line represents the STP-Net, the green line represents the base-Net. In the far right figure, the dotted line represents the loss of testing, the solid line represents the loss of training.}
\label{fig:semiNet}
\end{center}
\vskip -0.2in
\end{figure*}
Fig.~\ref{fig:dense} is the conventional fully-connected layer, where each orange circle represents an element in the input vector, each blue circle represents the neuron, and the line between orange circle and blue circle is a weight value. Fig.~\ref{fig:stp} is the STP layer, where lines with the same color represent the same weight value and each neuron can output $n$ values, here $n = 2$. Therefore, in STP layers, we just need $pq$ parameters which are $\frac{1}{n^2}$ of the number of parameters in dense layers, when the sizes of input and output are the same.

We train two small neural networks with two hidden layers. The one consists of dense layers, namely base-Net, and the other consists of STP-layers, namely STP-Net. Both Nets are used to approximate a periodic function $y=\sin(0.5\pi x)+\sin(\pi x)+\sin(2\pi x)$. As shown in Fig. \ref{fig:semiNet}, the generation capability of the STP-Net is better than the base-Net. 

\section{STP-tensor decomposition}
\label{sec:3}
\subsection{STTu decomposition }
\begin{figure}[ht]
\vskip 0.1in
\begin{center}
\includegraphics[width=0.5\columnwidth]{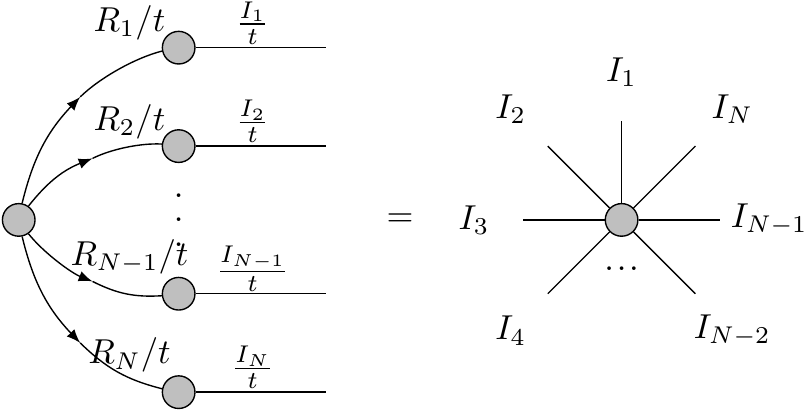} 
\caption{Tensor diagram of STTu decomposition.}
\label{fig:STTu}
\end{center}
\vskip -0.2in
\end{figure}

Tucker decomposition \cite{tucker1963implications} decomposes a tensor into a core tensor multiplied by a matrix along each mode, which can be denoted by
\begin{equation}
\begin{split}
\mathcal{X}&=\mathcal{G}\times_{1}\mathbf{A}^{(1)}\times_{2}\mathbf{A}^{(2)}\cdots\times_{N}\mathbf{A}^{(N)}\\
                       &=[\![\mathcal{G};\mathbf{A}^{(1)},\mathbf{A}^{(2)},\cdots,\mathbf{A}^{(N)}]\!],
\end{split}
\end{equation}
where $\mathcal{X} \in \mathbb{R}^{I_1\times I_2\times \cdots \times I_N}$ is the original tensor, $\mathcal{G}\in \mathbb{R}^{R_1 \times R_2 \times \cdots \times R_N}$ is the core tensor, and $\mathbf{A}^{(n)} \in \mathbb{R}^{I_n \times R_n}, n = 1, \cdots, N$ are factor matrices. We apply STP to Tucker decomposition to obtain the STTu decomposition, which can be denoted by
\begin{equation}
\begin{split}
\mathcal{X}&=\hat{\mathcal{G}}\ltimes_{1}\hat{\mathbf{A}}^{(1)}\ltimes_{2}\hat{\mathbf{A}}^{(2)}\cdots\ltimes_{N}\hat{\mathbf{A}}^{(N)}\\
                       &=[\![\hat{\mathcal{G}};\hat{\mathbf{A}}^{(1)},\hat{\mathbf{A}}^{(2)},\cdots,\hat{\mathbf{A}}^{(N)}]\!],
\end{split}
\end{equation}
where $\ltimes_k$ is the left semi-tensor mode-$n$ product, $\mathcal{X} \in \mathbb{R}^{I_1\times I_2\times \cdots \times I_N}$, $\hat{\mathcal{G}}\in \mathbb{R}^{R_1\times R_2\times \cdots \times R_N}$ is the core tensor, and $\hat{\mathbf{A}}^{(n)}\in\mathbb{R}^{ \frac{R_n}{t}\times \frac{I_n}{t}}, n=1,\cdots,N$ are factor matrices. Its tensor diagram notation is shown in Fig. \ref{fig:STTu}.

The left semi-tensor mode-$n$ product of $\hat{\mathcal{G}}$ and $\hat{\mathbf{A}}^{(n)}$ is formulated as
\begin{equation}
\bar{\mathcal{X}}=\hat{\mathcal{G}}\ltimes_{n}\hat{\mathbf{A}}^{(n)},
\end{equation}
which can be divided into three steps: 1) unfolding the tensor $\hat{\mathcal{G}}$ to the matricization representation $\hat{\mathbf{G}}_{(n)}\in \mathbb{R}^{R_n\times(R_1\times\cdots R_{n-1}R_{n+1}\cdots\times R_N)}$, 2) $\bar{\mathbf{X}}=\hat{\mathbf{G}}_{(n)}^\mathrm{T}\ltimes\hat{\mathbf{A}}^{(n)} \in \mathbb{R}^{(R_1\times\cdots R_{n-1}R{n+1}\cdots\times R_N)\times I_n}$, and 3) folding  $\bar{\mathbf{X}}$ to a tensor $\bar{\mathcal{X}} \in \mathbb{R}^{R_1\times\cdots\times R_{n-1} \times I_n\times R_{n+1} \times \cdots \times R_N}$

The memory ratio between Tucker decomposition and STTu decomposition is
\begin{equation}
 \frac{\prod_{n=1}^N R_n + \sum_{n=1}^N I_n R_n}{\prod_{n=1}^N R_n + \frac{1}{t^2}\sum_{n=1}^N I_n R_n}.
\end{equation}
 When $I_n\gg R_n$, the hyper-parameter $t$ can play a decisive role in the memory ratio between Tucker decomposition and STTu decomposition, but at the same time, the representation capability of the low-rank tensor will decline much. In addition, the number of parameters in Tucker decomposition scale exponentially with the tensor orders. 

\begin{figure*}[ht]
\vskip -0.1in
\begin{center}
\subfigure[a TR layer]{
\label{fig:TR layer} 
\includegraphics[width=0.45\columnwidth]{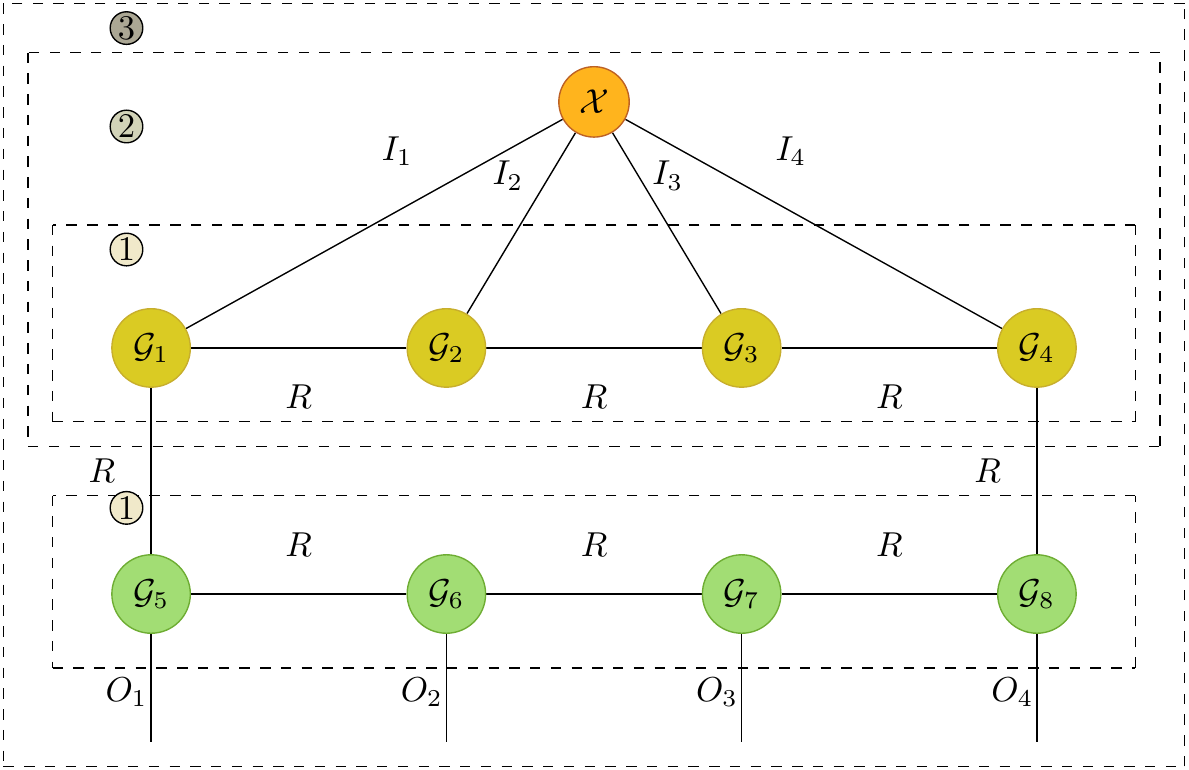}}
\subfigure[an STR layer]{
\label{fig:STR layer} 
\includegraphics[width=0.45\columnwidth]{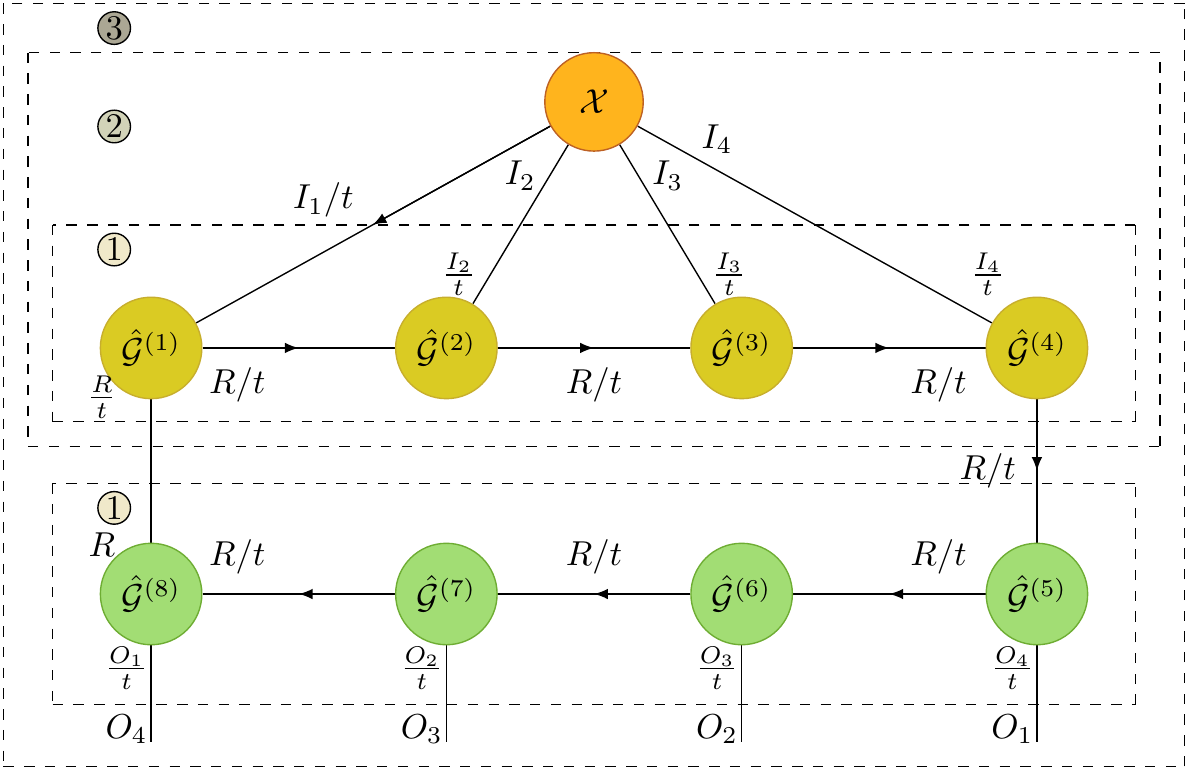}}

\subfigure[a TT layer]{
\label{fig:TT layer}
\includegraphics[width=0.45\columnwidth]{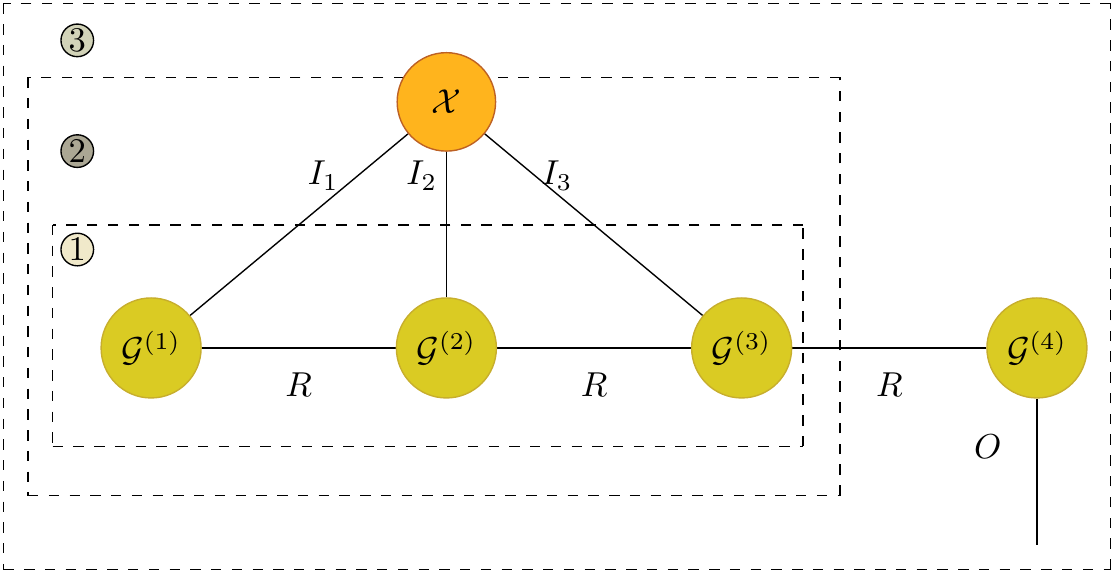}}
\subfigure[an STT layer]{
\label{fig:STT layer}
\includegraphics[width=0.45\columnwidth]{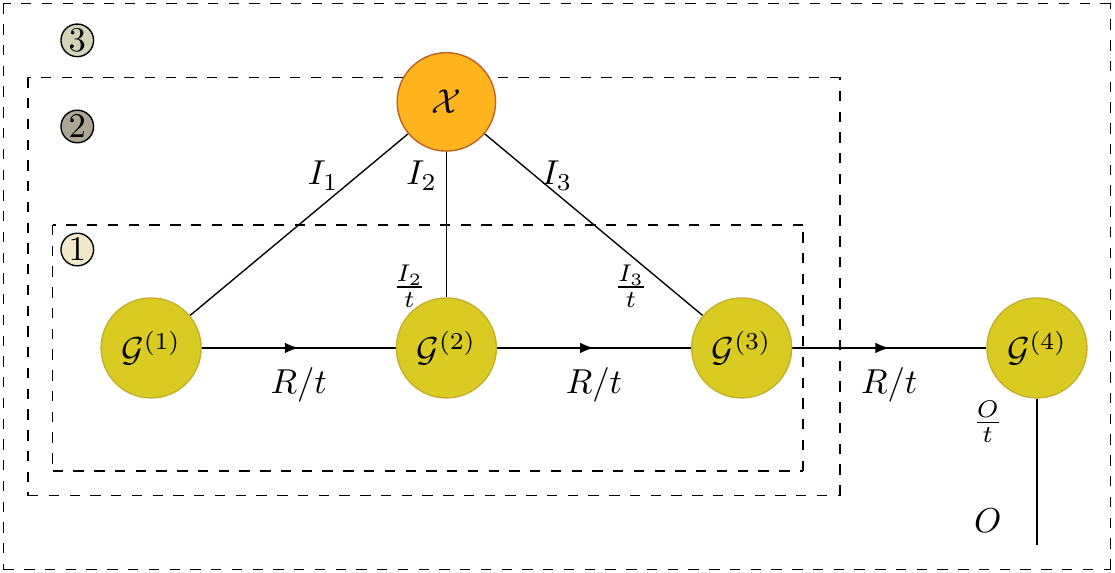}}
\label{fig:FCL}
\caption{Tensor diagram of FCL in TR-format, STR-format, TT-format and STT-format.}
\end{center}
\vskip -0.2in
\end{figure*}
\subsection{STR decomposition}
Both of TT and TR alleviate the problem of dimensionality curse. Compared with TT decomposition, TR decomposition does not highly depend on permutations of tensor dimensions, due to its cyclic structure. In this section, we focus on TR decomposition \cite{zhao2016tensor}, which provides an efficient representation for a high-dimensional tensor by a sequence of three-dimensional cores. Vectors and matrices have specific TT representations in neural networks \cite{novikov2015tensorizing}, thus we directly introduce the STT decomposition in network compression in Section \ref{sec:4}.

TR decomposition is denoted by 
\begin{equation}
\mathcal{X}=\sum^{R_1,\cdots,R_N}_{r_1,\cdots,r_N=1}\mathcal{G}^{(1)}(r_1,:,r_2)\circ \cdots \circ \mathcal{G}^{(N)}(r_N,:,r_{N+1}),
\end{equation}
where $\mathcal{X} \in \mathbb{R}^{I_1\times I_2\times \cdots \times I_N}$ is the original tensor, $r_{N+1}=r_1$, the symbol `$\circ$’ denotes the outer product of vectors, and $\mathcal{G}^{(n)}(r_n,:,r_{n+1})\in \mathbb{R}^{I_n}$ denotes the $\{r_n,r_{n+1}\}$-th mode-2 fiber of tensor factors $\mathcal{G}^{(n)}\in \mathbb{R}^{R_n\times I_n\times R_{n+1}}$. For simplicity, we assume $R_1=R_2=\cdots=R_N=R$, which is the size of tensor ring rank, and define $\operatorname{TR}$($\mathcal{X}$; $R$, $N$) as the operation to obtain $N$ factors $\mathcal{G}^{(n)}$ with tensor ring rank $R$ from $\mathcal{X}$. 

The merging process of two tensor ring cores is denoted by
\begin{equation}
\mathcal{G}^{(n,n+1)}(r_n,:,:,r_{n+2}) = \mathcal{G}^{(n)}(r_n,:,:) \mathcal{G}^{(n+1)}(:,:,r_{n+2}),
\label{equ:TR merge}
\end{equation}
where $\mathcal{G}^{(n,n+1)}\in \mathbb{R}^{R\times I_n\times I_{n+1}\times R}$. Replacing the matrix product by semi-tensor product, TR decomposition evolves to semi-tensor ring (STR) decomposition.

In STR decomposition, each tensor core is $\hat{\mathcal{G}}^{(n)}\in \mathbb{R}^{\frac{R}{t}\times \frac{I_n}{t}\times R}$, and the merging process of two tensor cores is denoted by
\begin{equation}
\hat{\mathcal{G}}^{(n,n+1)}(r_n,:,:,r_{n+2}) = \hat{\mathcal{G}}^{(n)}(r_n,:,:) \ltimes\hat{\mathcal{G}}^{(n+1)}(:,:,r_{n+2}),
\label{equ:STR merge}
\end{equation}
where $\hat{\mathcal{G}}^{(n,n+1)}(r_n,:,:,r_{n+2})\in \mathbb{R}^{\frac{I_n}{t}\times I_{n+1}}$ and $\hat{\mathcal{G}}^{(n,n+1)}\in \mathbb{R}^{\frac{R}{t}\times \frac{I_n}{t}\times I_{n+1}\times R}$.

The above operation can be represented in tensor form as 
\begin{equation}
\hat{\mathcal{G}}^{(n,n+1)} = \hat{\mathcal{G}}^{(n)}\ltimes_{3,1} \hat{\mathcal{G}}^{(n+1)},
\label{equ:STR merge2}
\end{equation}
where $\ltimes_{3,1}$ represents the STP between the third mode of the former tensor and the first mode of the latter tensor. Similar to definition $\operatorname{TR}$($\mathcal{X}$; $R$, $N$), we define $\operatorname{STR}$($\mathcal{X}$; $R$, $N$, $t$) as the operation to obtain $N$ factors $\hat{\mathcal{G}}^{(n)} \in \mathbb{R}^{\frac{R}{t}\times \frac{I_n}{t}\times R}$ with semi-tensor ring rank $R$ and semi-tensor product ratio $t$ from $\mathcal{X}$. 

The ratio of the memory between TR decomposition and STR decomposition is ${t^2}$.  Only the hyper-parameter $t$ affects the memory ratio between TR-Nets and STR-Nets.

\section{STP-Nets}
\label{sec:4}
STP-Nets are TNNs using STP-tensor decomposition. The forward pass in tensor neural networks do not merge the decomposed weight tensor firstly, while adopting the order with less computational complexity. For example, the convolution forward pass with Tucker-2 decomposition \cite{tucker1966some} is considered as three steps: 1) reducing the number of channels using the third-mode factor matrix, 2) doing regular convolution using the core tensor, 3) getting back to the output channels like the original convolution using the forth-mode factor matrix. Both  step 1 and  step 3 take mode-3 product. Therefore, in the convolution forward pass with STTu decomposition, the mode-3 product is replaced by the semi-tensor mode-3 product.

In Section \ref{sec:3}, we have given the expression of STTu decomposition and STR decomposition. In this section, we will introduce STT-Nets and STR-Nets in details from fully-connected layer (FCL) and convolutional layer (ConvL) aspects. We call FCL which compressed by TR (or TT) decomposition and STR (or STT) decomposition as TR (or TT) layer and STR (or STT) layer, respectively, and call corresponding ConvL as TR-conv (or TT-conv) layer and STR-conv (or STT-conv) layer.

\begin{table*}[t]
\caption{The formulas of every steps in the STR layer and the flops of every steps in the STR layer and the TR layer. $B$ is the batch size of testing samples}
\label{tab:STR-FCL}
\begin{center}
\begin{small}
\begin{tabular}{lccc}
\toprule
   & STR formula & STR flops & TR flops\\
\midrule
step (1) & \tabincell{c}{ $\hat{\mathcal{G}}_{in} = \operatorname{Merge1} \{\hat{\mathcal{G}}^{(1)-(N)}\}$\\$\hat{\mathcal{G}}_{out} = \operatorname{Merge1} \{\hat{\mathcal{G}}^{(N)-(N+\hat{N})}\}$} &$\le\frac{4}{t^3}R^3(I+O)$  &  $\le 4R^3(I+O)$   \\
step (2) & $\mathbf{Y}=\langle{\mathcal{X}\ltimes_{1,2}\hat{\mathcal{G}_{in}}}\rangle_{N-1}\in \mathbb{R}^{R\times R}$  & $\frac{2}{t}BR^2I$   &   $2BR^2I$  \\
step (3) & $\mathcal{Y}=\langle{\mathbf{Y}\ltimes_{2,1}\hat{\mathcal{G}}_{out}}\rangle_1 \in \mathbb{R}^{O_1\times\cdots\times O_{\hat{N}}}$  & $\frac{2}{t}BR^2O$  &   $2BR^2O$   \\
\midrule
Total & -  & $\le(\frac{4}{t^3}R^3+\frac{2}{t}BR^2)(I+O)$ & $\le(4R^3+2BR^2)(I+O)$\\
\bottomrule
\end{tabular}
\end{small}
\end{center}
\end{table*}

\subsection{Fully-Connected Layer}
When the fully-connected layer is the middle layer, the weight matrix $\mathbf{W}$ in Eq. (\ref{equ:FCL}) is reshaped to the weight tensor $\mathcal{W}$ in Eq. (\ref{equ:TCL}). When the fully-connected layer is the last layer, the weight tensor is $\mathcal{W}^\prime\in \mathbb{R}^{O\times I_1 \times I_2 \times \cdots\times I_N}$. Whatever, the input is always the tensor $\mathcal{X}\in\mathbb{R}^{I_1\times I_2\times\cdots\times I_N}$. Because of the flexibility of TR decomposition, it is suitable for the two above situation. Novikov \etal define two different TT-representations \ie TT-matrix and TT-vector, corresponding to the above two situation, respectively \cite{novikov2015tensorizing}. The tensor diagram of TT-matrix is given in Fig.~\ref{fig:TTN}. The fully-connected layer is usually applied to the last layer of convolutional neural networks. Thus in this section, we introduce the TT-vector format and the STT-vector format.

\subsubsection{STR Layer}
TR layers are the tensor regression layers, which use  $\operatorname{TR}$($\mathcal{W}$; $R$, $N+\hat{N}$) to obtain $N+N^\prime$ tensor ring factors, $\{\mathcal{G}^{(n)}\}_{n=1}^N \in \mathbb{R}^{R\times I_n \times R}$ and $\{\mathcal{G}^{(n)}\}_{n=N+1}^{N+\hat{N}} \in \mathbb{R}^{R\times O_{n-N} \times R}$. The tensor diagram of a TR layer is given in Fig. \ref{fig:TR layer}, which shows  the computational order. The results in step (1) is obtained by using hierarchical merging, whose computational cost is cheaper than sequential merging when $N$ and $\hat{N}$ is large \cite{wang2018wide}. 

Similarly, STR layers use $\operatorname{STR}$($\mathcal{W}$; $R$, $N+\hat{N}$, $t$) to obtain $N+N^\prime$ semi-tensor ring factors, $\{\hat{\mathcal{G}}^{(n)}\}_{n=1}^N \in \mathbb{R}^{\frac{R}{t}\times \frac{I_n}{t} \times R}$ and $\{\hat{\mathcal{G}}^{(n)}\}_{n=N+1}^{N+\hat{N}} \in \mathbb{R}^{\frac{R}{t}\times \frac{O_{n-N}}{t} \times R}$, and the tensor diagram of an STR layer is given in Fig. \ref{fig:STR layer}. The merging result of $\{\hat{\mathcal{G}}^{(1)},\hat{\mathcal{G}}^{(2)},\cdots,\hat{\mathcal{G}}^{(N)}\}$ is formulated as
\begin{equation}
\operatorname{Merge1} \{\hat{\mathcal{G}}^{(1)-(N)}\} = \hat{\mathcal{G}}^{(1)}\ltimes_{3,1}\hat{\mathcal{G}}^{(2)}\ltimes_{4,1},\cdots,\ltimes_{N+1,1}\hat{\mathcal{G}}^{(N)} \in \mathbb{R}^{\frac{R}{t}\times \frac{I_1}{t}\times I_2\times\cdots\times I_{N-1}\times I_{N}\times R},\nonumber
\label{equ:merge1}
\end{equation}
which  hierarchically merges like TR decomposition does.

The computational order in an STR layer is the same as in a TR layer. Table \ref{tab:STR-FCL} gives the formulas and computation costs of every steps. Except the computation complexity of STR in step (1) is ${1}/{t^3}$ of TR, the computational complexity ratio of STR and TR is ${1}/{t}$. The batch size is usually set to 128, which is larger  than the rank size. Thus the computation complexity of STR layers is approximately ${1}/{t}$ that of TR layers. 

\subsubsection{STT Layer}
\begin{figure*}[ht]
\vskip -0.2in
\begin{center}
\subfigure[a TR-conv layer]{
\label{fig:TR-conv layer} 
\includegraphics[width=0.45\columnwidth]{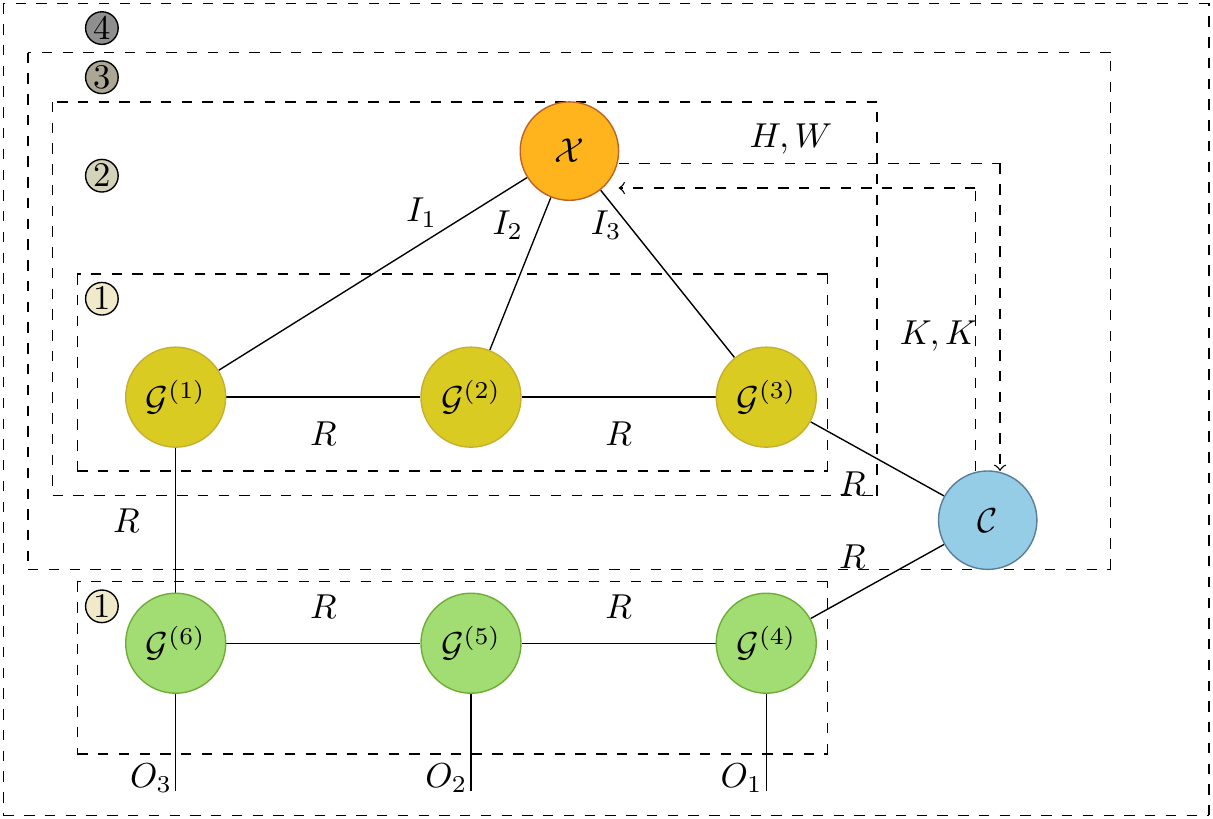}}
\subfigure[an STR-conv layer]{
\label{fig:STR-conv layer} 
\includegraphics[width=0.45\columnwidth]{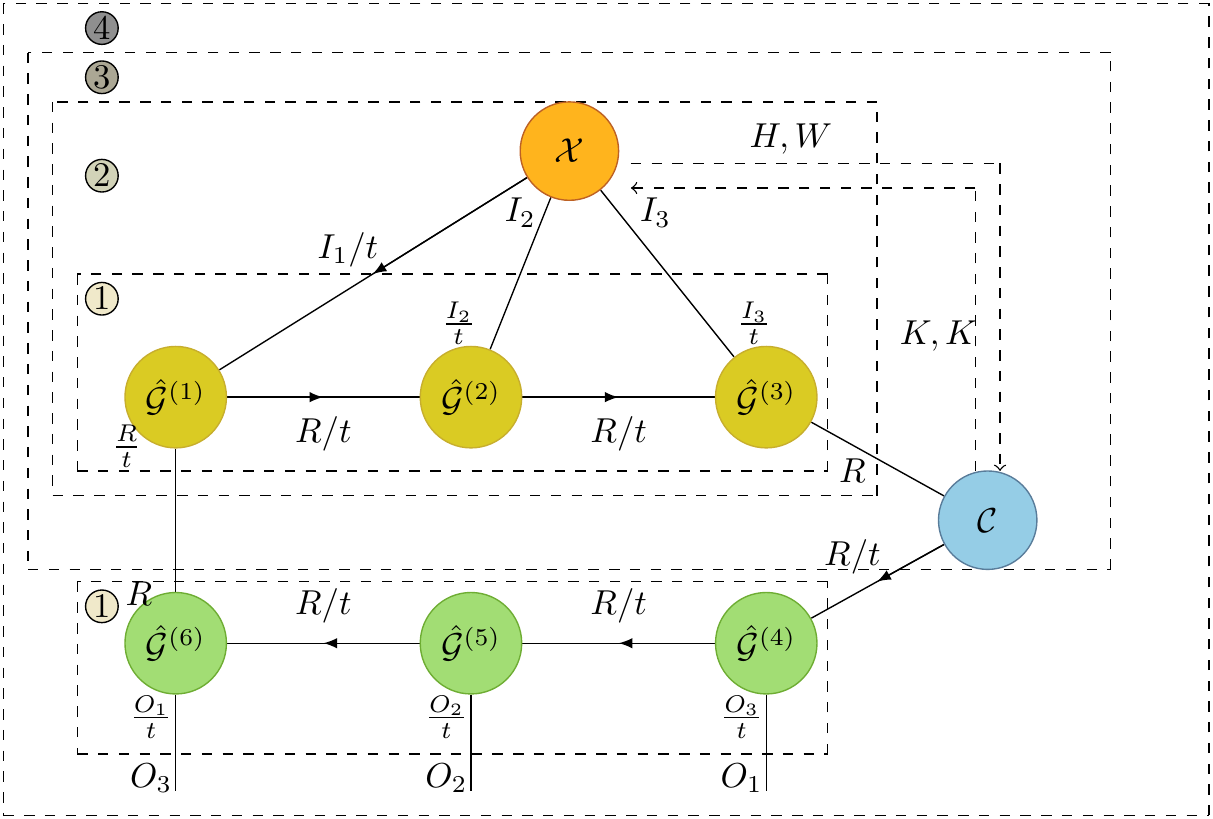}}

\subfigure[a TT-conv layer]{
\label{fig:TT-conv layer}
\includegraphics[width=0.45\columnwidth]{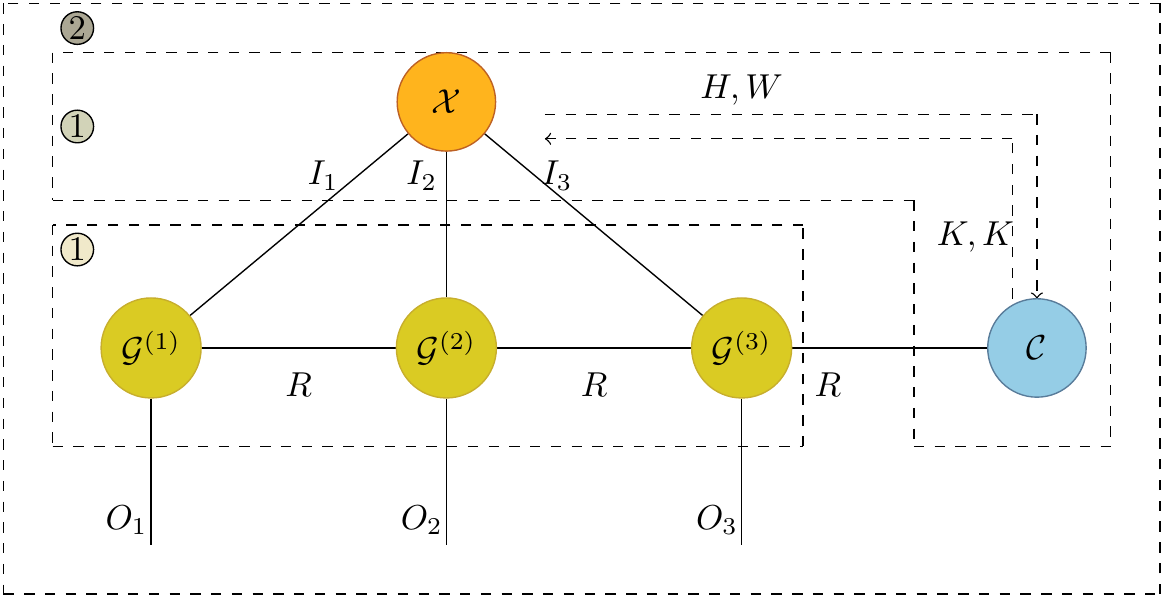}}
\subfigure[an STT-conv layer]{
\label{fig:STT-conv layer}
\includegraphics[width=0.45\columnwidth]{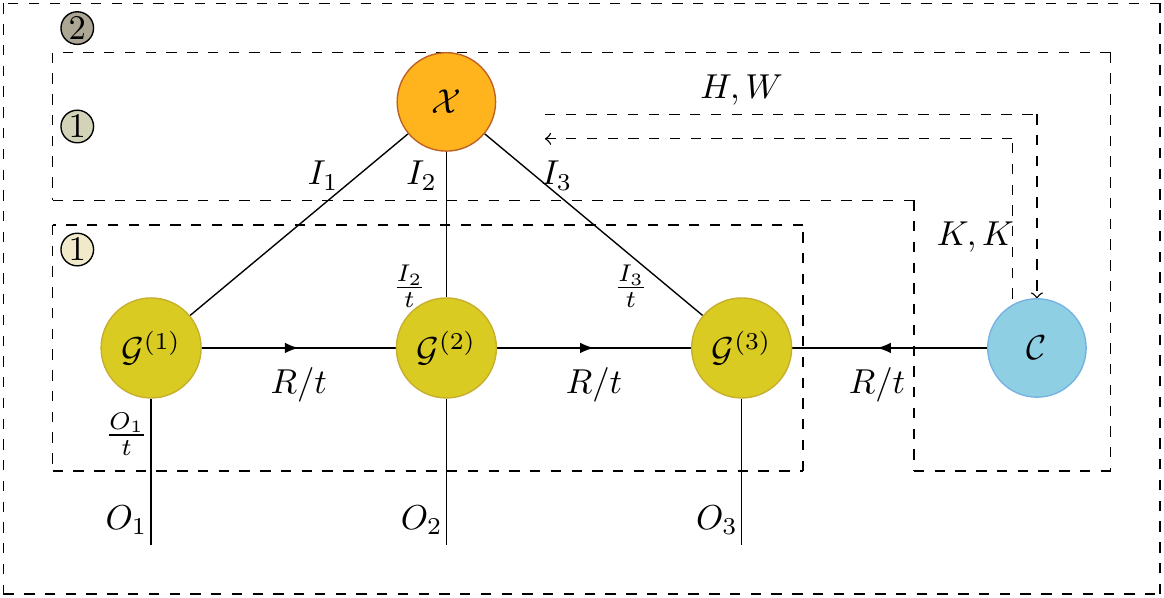}}
\caption{Tensor diagram of ConvL in TR-format,  STR-format, TT-format and STT-format.}
\label{fig:ConvL}
\end{center}
\vskip -0.2in
\end{figure*}
\begin{table*}[t]
\caption{The formulas of every steps in the STT layer and the flops of every steps in the STT layer and the TT layer. $B$ is the batch size of testing samples}
\label{tab:STT-FCL}
\begin{center}
\begin{small}
\begin{tabular}{lccc}
\toprule
   & STT formula & STT flops & TT flops\\
\midrule
step (1) & $\hat{\mathcal{U}}_{in} = \operatorname{Merge2} \{\hat{\mathcal{G}}^{(1)-(N)}\}$ &$\le\frac{4}{t}R^2I$  &  $\le 4R^2I$   \\
step (2) & $\mathbf{y}=\langle{\mathcal{X},\hat{\mathcal{U}_{in}}}\rangle_N \in \mathbb{R}^{R}$  & $2BRI$  & $2BRI$  \\
step (3) & $\mathcal{Y}=\mathbf{y}\ltimes\hat{\mathbf{U}_{N+1}} \in \mathbb{R}^{O}$  & $\frac{2}{t}BRO$  &   $2BRO$   \\
\midrule
Total & -  & $\le(\frac{4}{t}R^2+2BR)I+\frac{2}{t}BRO$ & $\le(4R^3+2BR)I+2BRO$\\
\bottomrule
\end{tabular}
\end{small}
\end{center}
\end{table*}
In a TT-vector layer, the intact tensor $\mathcal{W}^\prime$ is decomposed to $N+1$ TT-vector factors, $\{\mathcal{U}^{(n)}\}_{n=1}^{N}\in \mathbb{R}^{R_{n-1}\times I_n \times R_n}$ and $\mathcal{U}^{N+1}\in \mathbb{R}^{R_N\times O \times R_{N+1}}$, where $R_{0}=R_{N+1}=1$ and we set $R_1=R_2=\cdots=R_N=R$. The tensor diagram of the TT-vector layer is given in Fig.~\ref{fig:TT layer}, which shows the computational order.

Similarly, in the STT-vector layer, the intact tensor is decomposed to $N+1$ STT-vector factors, $\hat{\mathcal{U}}^{(1)} \in \mathbb{R}^{i\times I_n\times R}$, $\hat{\mathcal{U}}^{(n)}\in \mathbb{R}^{\frac{R}{t}\times \frac{I_n}{t} \times R}$, $ n = 2, \cdots, N $,  and $\hat{\mathbf{U}}^{N+1}\in \mathbb{R}^{\frac{R}{t}\times \frac{O}{t}}$, and the tensor diagram of an STT-vector layer is given in Fig. \ref{fig:STT layer}. The merging result of $\{\hat{\mathcal{U}}^{(1)},\hat{\mathcal{U}}^{(2)},\cdots,\hat{\mathcal{U}}^{(N)}\}$ is formulated as
\begin{equation}
\operatorname{Merge2} \{\hat{\mathcal{U}}^{(1)-(N)}\}  = \hat{\mathcal{U}}^{(1)}\ltimes_{2,1}\hat{\mathcal{U}}^{(2)}\ltimes_{3,1},\cdots,\ltimes_{N,1}\hat{\mathcal{U}}^{(N)} \in \mathbb{R}^{I_1\times I_2\times\cdots\times I_{N-1}\times I_{N}\times R}.\nonumber
\label{equ:merge2}
\end{equation}

The computational order in an STT-vector layer is the same as it in a TT-vector layer. Table \ref{tab:STT-FCL} gives the formulas and computation costs of every steps. The STT-vector layer and the TT-vector layer have the same Step (2). The computational complexity ratio of them in steps (1) and (3)  is $\frac{1}{t}$. The batch size is usually set to 128, which is larger  than the rank size, and $O$ is less than $I$. Thus the computation complexity of STT-vector layers is slightly less than that of TT-vector layers. 

The parameters of a TT-vector layer and an STT-vector layer are given in Table~\ref{tab:TT-number2}. Thus it can be seen that the number of parameters in the STT layer is about $\frac{1}{t^2}$ of that in the TT layer.

\begin{table}[ht]
    \centering
    \renewcommand\arraystretch{1.5}
    \caption{The compare of parameters between the TT layer and the STT layer.}
    \begin{tabular}{|c|c|c|}    
    \hline
                & TT layer & STT layer \\
        \hline
        Param. &  {$R(I_1+O)+\sum\limits_{n=2}^N R^2I_n$}& {$R I_1+\frac{1}{t^2}(R O_1 +\sum\limits_{n=2}^{N-1} R^2I_n)$} \\
        \hline
    \end{tabular}
    \label{tab:TT-number2}
\end{table}

\subsection{Convolutional Layer}
In a convolutional layer, an input tensor $\mathcal{X} \in \mathbb{R}^{H \times W \times I}$ is convoluted with a 4th order kernel tensor
$\mathcal{K} \in \mathbb{R}^{K \times K \times I \times O}$  and mapped to a 3rd order tensor $\mathcal{Y} \in \mathbb{R}^{H^\prime\times W^\prime\times O}$, the mathematical formulation is as follows
\begin{equation}
\begin{split}
\mathcal{Y}_{h^\prime,w^\prime,o} = \Sigma_{k_1,k_2=1}^K\Sigma_{i=1}^I \mathcal{X}_{h,w,i}\mathcal{K}_{k_1,k_2,i,o},
\label{equ:conv}
\end{split}
\end{equation}
where $h =(h^\prime-1)S + k_1 - P$, $w =(w^\prime-1)S + k_2 - P$, $S$ is stride size, $P$ is zero-padding size, $K$ is convolutional window size, $I$ and $O$ are the number of input and output channels, respectively. Eq. (\ref{equ:conv}) is abbreviated in tensor form as $\mathcal{Y} = \mathcal{X} \star \mathcal{K}$, where $\star$ represents convolutional operation.

\begin{table*}[t]
\caption{The formulas of every steps and their flops in the STR-conv layer. $B$ is the batch size of testing samples}
\label{tab:STR-ConvL}
\begin{center}
\begin{small}
\begin{tabular}{lccc}
\toprule
   & STR-conv formula & STR-conv flops & TR-conv flops\\
\midrule
step (1) & \tabincell{c}{ $\hat{\mathcal{G}}_{in} = \operatorname{Merge} \{\hat{\mathcal{G}}^{(1)-(N)}\}$\\$\hat{\mathcal{G}}_{out} = \operatorname{Merge} \{\hat{\mathcal{G}}^{(N)-(N+\hat{N})}\}$} &$\le\frac{4}{t^3}R^3(I+O)$  &  $\le 4R^3(I+O)$   \\
step (2) & $\mathcal{Y}_1=\langle{\mathcal{X}\ltimes_{3,2}\hat{\mathcal{G}_{\rm in}}}\rangle_{N-1}\in \mathbb{R}^{H\times W \times R\times R}$  & $\frac{2}{t}BHWR^2I$   &   $2BHWR^2I$  \\
step (3) & $\mathcal{Y}_2(:,:,r,:) = \mathcal{Y}_1(:,:,r,:) \star \mathcal{C} \in \mathbb{R}^{H^\prime\times W^\prime \times R}$  & $2H^\prime W^\prime R^2 K^2$  &   $2H^\prime W^\prime R^2 K^2$   \\
step (4) &$\mathcal{Y}=\langle{\mathcal{Y}_2 \ltimes_{4,1}\hat{\mathcal{G}_{\rm out}}}\rangle_{1}\in \mathbb{R}^{H^\prime\times W^\prime \times O_1\times \cdots \times O_{\hat{N}}}$ &$\frac{2}{t}BH^\prime W^\prime R^2O$ &$2BH^\prime W^\prime R^2O$\\
\midrule
Total & -  & \tabincell{c}{$\le(\frac{4}{t^3}R+\frac{2}{t}BHW)*$\\$R^2(I+O)+2HWR^2K^2$}& \tabincell{c}{$\le(4R+2BHW)*$\\$R^2(I+O)+2HWR^2K^2$}\\
\bottomrule
\end{tabular}
\end{small}
\end{center}
\end{table*}
\begin{table*}[t]
\caption{The formulas of every steps and their flops in the STT-conv layer. $B$ is the batch size of testing samples}
\label{tab:STT-ConvL}
\begin{center}
\begin{small}
\begin{tabular}{lccc}
\toprule
   & STT-conv formula & STT-conv flops & TT-conv flops\\
\midrule
step (1) & \tabincell{c}{$\hat{\mathcal{V}} = \operatorname{Merge3} \{\hat{\mathcal{V}}^{(1)-(N)}\}$\\$\mathcal{Y}_1(:,:,:,i_1,\cdots,i_N) = \mathcal{X}(:,:,:,i_1,\cdots,i_N)\star\mathcal{C}\in\mathbb{R}^{2H^\prime \times W^\prime\times R}$} &\tabincell{c}{ $\le\frac{4}{t^2}R^2IO$\\$2H^\prime W^\prime R K^2$}  & \tabincell{c}{$\le 4R^2IO$\\$2H^\prime W^\prime R K^2$}   \\
step (2) &$\mathcal{Y}_2=\langle{\mathcal{Y}_1 \ltimes_{3,1}\hat{\mathcal{V}}}\rangle_{N}\in \mathbb{R}^{H^\prime\times W^\prime \times O_1\times \cdots \times O_{\hat{N}}}$ &$\frac{2}{t}B H^\prime W^\prime IOR$ &$2B H^\prime W^\prime IOR$\\
\midrule
Total & -  & \tabincell{c}{$\le\frac{2}{t}IOR(\frac{2}{t}R+BHW)$\\$+2HWR^2K^2$}& \tabincell{c}{$\le2IOR(2R+BHW)$\\$+2HWR^2K^2$}\\
\bottomrule
\end{tabular}
\end{small}
\end{center}
\end{table*}
\subsubsection{STR-conv Layer}
In a TR-conv layer, $I$ and $O$ are reshaped to the size $I_1 \times I_2 \times \cdots \times I_N$ and the size $O_1 \times O_2 \times \cdots \times O_{\hat{N}}$, respectively, while $D$ is unchanged in the kernel tensor, because $I$ and $O$ are much larger than $K$ and the spatial information can be maintained. Thus the size of the input tensor $\mathcal{X}$ is $H \times W \times I_1\times I_2\times\cdots\times I_N$. The kernel tensor $\mathcal{K}$ is decomposed to $N+\hat{N}$ TR factors $\mathcal{G}^{(1)},\cdots,\mathcal{G}^{(N)},\mathcal{G}^{(N+1)},\cdots,\mathcal{G}^{(N+\hat{N})}$ and a convolutional kernel $\mathcal{C}\in \mathbb{R}^{K\times K\times R\times R}$. Its tensor diagram is given in Fig. \ref{fig:TR-conv layer}, which shows the computational order.

Differently, in an STR-conv layer, the kernel tensor $\mathcal{K}$ is decomposed to $N+\hat{N}$ STR factors $\hat{\mathcal{G}}^{(1)},\cdots,\hat{\mathcal{G}}^{(N)},\hat{\mathcal{G}}^{(N+1)},\cdots,\hat{\mathcal{G}}^{(N+\hat{N})}$, but the convolutional kernel $\mathcal{C}\in \mathbb{R}^{K\times K\times R\times R}$ is the same. And its tensor diagram is given in Fig. \ref{fig:STR-conv layer}. 

The computational order in an STR-conv layer is the same as in a TR-conv layer. Table \ref{tab:STR-ConvL} gives the formulas and computation costs of every steps. The batch size is usually set to 128, which is larger far than the rank size. Thus  the computation complexity of STR layers is approximately ${1}/{t}$   of TR layers. The computational complexity of semi-tensor product operation is less than that of matrix product, thus the training and testing time of STR-Nets is less than TR-Nets.

\subsubsection{STT-conv Layer}
Garipov \etal apply TT-matrix in convolutional layers \cite{garipov2016ultimate}. In a TT-conv layer, $I$ and $O$ are reshaped to the size $I_1 \times I_2 \times \cdots \times I_N$ and the size $O_1 \times O_2 \times \cdots \times O_{\hat{N}}$, respectively. Different from a TR-conv layer, $\hat{N}$ has to be the same as $N$ in a TT-conv layer and the size of the input tensor $\mathcal{X}$ is $H \times W \times 1 \times I_1 \times I_2 \times \cdots \times I_N$. The kernel tensor $\mathcal{K}\in \mathbb{R}^{I_1\times \cdots \times I_N \times O_1\times \cdots \times O_N \times K\times K}$ is decomposed to $N+1$ TT-matrix factors $\{\mathcal{V}^{(n)}\}_{n=1}^{N} \in \mathbb{R}^{R_{n-1}\times I_n\times O_n\times R_{n}}$ and $\mathcal{C}\in \mathbb{R}^{K\times K\times R_{N+1}\times R_N}$, where $R_0=R_{N+1}=1$ and $\mathcal{C}$ is a small convolutional kernel. Here, we set $R_1=R_2=\cdots=R_{N} = R$. Its tensor diagram is given in Fig.~\ref{fig:TT-conv layer}, which shows the computational order.

In an STT-conv layer, the convolutional kernel tensor $\mathcal{K}\in \mathbb{R}^{I_1\times \cdots \times I_N \times O_1\times \cdots \times O_N \times K\times K}$ is decomposed to $N$ STT-matrix factors $\hat{\mathcal{V}}^{(1)},\hat{\mathcal{V}}^{(2)},\cdots,\hat{\mathcal{V}}^{(N)}$ and a small convolutional kernel $\mathcal{C}\in\mathbb{R}^{K\times K\times 1\times R}$, which is the same as that in the TT-conv layer. The structure of STT factors is more complex than STR factors, where  $\hat{\mathcal{V}}^{(1)}\in \mathbb{R}^{I_1\times \frac{O_1}{t}\times R}$, $\{\hat{\mathcal{V}}^{(n)}\}_{n=1}^{N}\in \mathbb{R}^{\frac{R}{t}\times \frac{I_n}{t}\times O_n\times R}$, $\hat{\mathcal{V}}^{(N)}\in \mathbb{R}^{\frac{R}{t}\times \frac{I_N}{t}\times O_N\times \frac{R}{t}}$. The merging result of $\{\hat{\mathcal{V}}^{(1)},\hat{\mathcal{V}}^{(2)},\cdots,\hat{\mathcal{V}}^{(N)}\}$ is formulated as
\begin{equation}
\begin{split}
\operatorname{Merge3} \{\hat{\mathcal{V}}^{(1)-(N)}\}  &= \hat{\mathcal{V}}^{(1)}\ltimes_{3,1}\hat{\mathcal{V}}^{(2)}\ltimes_{5,1},\cdots,\ltimes_{2N-1,1}\hat{\mathcal{V}}^{(N)} \\
 &\in \mathbb{R}^{\frac{R}{r}\times I_1\times \cdots\times I_{N}\times \frac{O_1}{t}\times \cdots\times O_{N}}.\nonumber
\end{split}
\label{equ:merge3}
\end{equation}

The computational order in an STT-conv layer is the same as in a TT-conv layer. Tab. \ref{tab:STT-ConvL} gives the formulas and computation costs of every steps. The convolutional operation in step (1) of the STT-conv and the TT-conv layer is the same. The computation complexity ratio of them in merging operation of step (1) is $\frac{1}{t^2}$ and that in step (2) is $\frac{1}{t}$. The batch size is usually set to 128, which is larger far than the rank size, and $IO$ is lager than $K^2$. Thus the computation complexity of STT-conv layers is about $\frac{1}{t}$ of that of TT-conv layers.

The parameters of a TT-conv layer and an STT-conv layer are given in Table~\ref{tab:TT-number}. Obviously, parameters in an STT-conv layer are less than a TT-conv layer.

\begin{table}[ht]
    \centering
    \renewcommand\arraystretch{1.5}    \caption{The compare of parameters between the TT-conv layer and the STT-conv layer.}
    \begin{tabular}{|c|c|c|}
    \hline
                & TT-conv layer & STT-conv layer \\
        \hline
        Param. &  \tabincell{c}{$R(I_1O_1+K^2)+$\\$ \sum_{n=2}^N R^2I_n O_n$}& \tabincell{c}{$R(\frac{1}{t}I_1O_1+K^2)+$\\$ \frac{1}{t^2}\sum_{n=2}^{N-1} R^2I_n O_n+\frac{1}{t^3}R^2I_N O_N$} \\
        \hline
    \end{tabular}
    \label{tab:TT-number}
\end{table}

\section{Experiments}
\label{sec:5}
In this section, we design STTu-Nets, STT-Nets and STR-Nets to verify their effectiveness in model compression. Because of the rise of convolutional neural networks, we focus on LeNet-5 \cite{lecun1998gradient}, and ResNet \cite{He_2016_CVPR} and WideResNet, where the former is relatively small and the latter are up to dozens of layers. WideResNet is wider than ResNet ten factors. In our proposed STP-Net, hyper-parameter $t$ also decides the compression rate, and we let $t$ be 2 for the following experiments. The reason is that the dimension of each order is small enough and each dimension needs to be a multiple of $t$.

\textbf{Datasets.} The simple network is used to classify MNIST which is a simple dataset for the task of handwritten digit recognition, composed of  60000  $28\times 28$ grey scale images for training and 10000 for testing. The deeper network is applied in Cifar10 \cite{krizhevsky2009learning} dataset which is more complex than MNIST. MINIST consists of $32 \times 32$ 3-channel image assigned to 10 different classes: airplane, automobile, bird, cat, deer, dog, frog, horse, ship, truck, and containing 50000 training and 10000 testing images. Besides of Cifar10, we also use Cifar100 for verifying the deeper network. 

\textbf{Training Details.} For the above datasets, validation images which are separated from training datasets are $10\%$  of training datasets, and the results come from the testing datasets using the best model for the validation datasets. The training settings of the learning rate, batch size, optimization, and epochs are not the same in both kinds of networks and the details will be introduced in the corresponding subsection. All the networks are trained using Pytorch \cite{paszke2019pytorch} and all the experiments are implemented on Nvidia GTX 1080 Ti GPUs.

\textbf{Assessment Criteria.} One evaluation metric is compression factor (CF), which is defined as the ratio between the number of parameters in the original network and that in the compressed network. The other is compression ratio, and it is defined as the ratio between  the number of parameters in the compressed network and that in the original network.

\subsection{Simple Network}

\begin{figure}[t]
\vskip -0.2in
\begin{center}
\includegraphics[width=0.5\columnwidth]{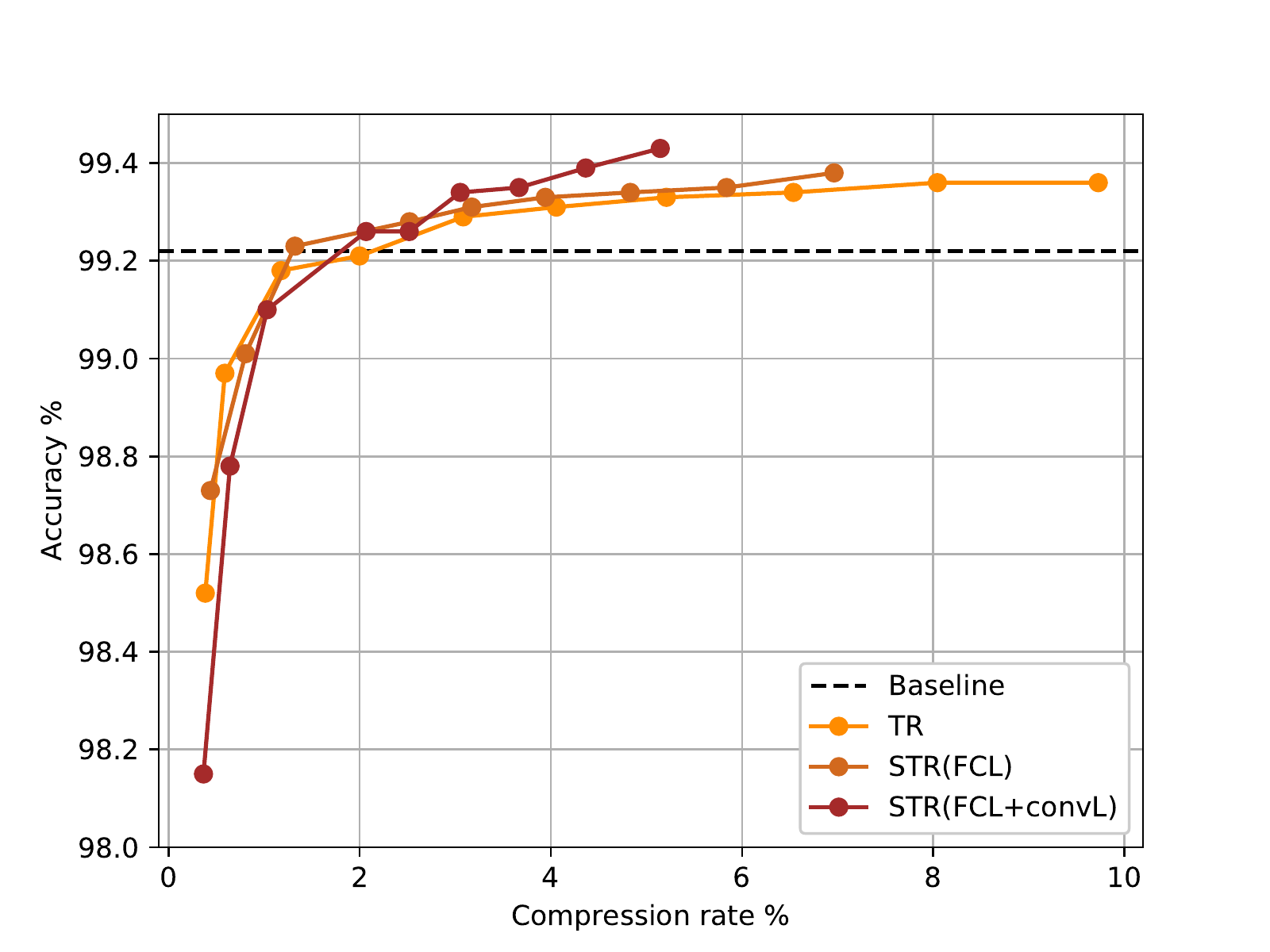} 
\caption{Accuracy vs. Model Compression Ratio for LeNet-5 on MNIST classification. The rank ranges from 4 to 20, with an interval of 2.}
\label{fig:cr_lenet5}
\end{center}
\vskip -0.2in
\end{figure}

We first investigate the effectiveness of our proposed models in a small network. LeNet-5 is a convolutional neural network with 2 convolution layers and 2 fully-connected layers. The dimensions in the STR-Nets based on LeNet-5 are similar to the TR-Nets in \cite{wang2018wide}), if one dimension of the channels is odd, it pluses one. 

Here, we give two kinds of situations. One is that we only compress FCL using STR decomposition, and the other is to compress both FCL and ConvL at the same time. As shown in Fig.~\ref{fig:cr_lenet5}, the STR-Nets get a higher compression factor than TR-Nets. When the rank is large (rank = 20), the accuracy of the STR-Net even is higher than the TR-Net, as shown in Table \ref{tab:LeNet results}.

\begin{table}[ht]
\caption{The simple network results. LeNet-5 on MNIST dataset. Testing time is per 10000 samples.}
\label{tab:LeNet results}
\begin{center}
\begin{tabular}{|l|c|c|c|c|}
\hline
Method & Params & CF & Acc\%&Test(s)\\
\hline\hline
LeNet-5& 429K & 1$\times$ &99.22&1.2\\
\hline
TR($R$=20) \cite{wang2018wide}& 41.7K &$10\times$&99.34&1.5\\
STR(FCL, $R$=20) &29.9K &$14\times$ &99.38 &1.6  \\
STR(FCL+ConvL, $R$=20) &22.0K &$\textbf{19}\times$ &\textbf{99.43}&1.5  \\
\hline
\end{tabular}
\end{center}
\end{table}

\textbf{Training settings.} The learning rates are 10 times in the first 10 epochs than ones in \cite{wang2018wide}. Other settings are the same. The initialization is in Gaussian distribution and the variance set has been analyzed in \cite{wang2018wide}.

\subsection{Deep Network}
\begin{table*}
\caption{The deep network compression.}
\begin{center}
\begin{tabular}{|l|c|c|c|c|c|c|}
\hline
        & \multicolumn{2}{c|}{uncompressed dimensions} & \multicolumn{2}{c|}{TR-Nets dimensions}& \multicolumn{2}{c|}{STR-Nets dimensions}\\
\hline\hline
layer & shape & Param. & shape of composite tensor & Param. & shape of composite tensor & Param. \\
\hline
ConvL1 & $3\times 3\times 3\times 16 $  & 432 & $9\times 3\times (4\times 2 \times 2)$&$20R^2$& $9\times 3\times (4\times 4)$&$14R^2$\\
\hline
unit1 & \tabincell{c}{ResBlock(3,16,16)\\ ResBlock(3,16,16) $\times$ 4} & \tabincell{c}{4608\\18432}& \tabincell{c}{$9\times (4\times 2\times 2 )\times (4\times 2\times 2)$\\$9\times (4\times 2\times 2)\times (4\times 2\times 2)$} &  \tabincell{c}{$50R^2$\\$200R^2$}& \tabincell{c}{$9\times (4\times 4)\times (4\times 4)$\\$9\times (4\times 4)\times (4\times 4)$} &  \tabincell{c}{$26R^2$\\$104R^2$}\\
\hline
unit2 & \tabincell{c}{ResBlock(3,16,32)\\ ResBlock(3,32,32) $\times$ 4} & \tabincell{c}{13824\\73728} &\tabincell{c}{$9\times (4\times 2\times 2)\times (4\times 4\times 2)$\\$9\times (4\times 4\times 2)\times (4\times 4\times 2)$} &   \tabincell{c}{$56R^2$\\$232R^2$}&\tabincell{c}{$9\times (4\times 4)\times (4\times 8)$\\$9\times (4\times 8)\times (4\times 8)$} &   \tabincell{c}{$28R^2$\\$120R^2$}\\
\hline
unit3 & \tabincell{c}{ResBlock(3,32,64)\\ ResBlock(3,64,64) $\times$ 4} & \tabincell{c}{55296\\294912}& \tabincell{c}{$9\times (4\times 4\times 4)\times (4\times 4\times 4)$\\$9\times (8\times 8)\times (4\times 4\times 4)$}  &   \tabincell{c}{$64R^2$\\$264R^2$}& \tabincell{c}{$9\times (4\times 8)\times (8\times 8)$\\$9\times (8\times 8)\times (8\times 8)$}  &   \tabincell{c}{$32R^2$\\$136R^2$}\\
\hline
FCL1 & $64\times 10 $ & 650 & $(4\times 4\times 4)\times 10  $ &  $22R^2$& $(4\times 4\times 4)\times 10  $ &  $5.5R^2$\\
\hline
Total & - &0.46M&  - &$908R^2$&  - &$465.5R^2$\\
\hline
\end{tabular}
\end{center}
\label{tab:resnet dimension}
\end{table*}

In general, deeper neural networks can handle more complex tasks, and we further evaluate the performance of STP-Nets on the Cifar10 image classification task, which is a more difficult classification task, based on ResNet. ResNet can be set up to 1202 layers, but we just use ResNet-32 and its dimensions are given in Table \ref{tab:resnet dimension}. Here, TR decomposition-based ResNet-32 is called TR-RNs, where using 's' is because of existing different rank size. The other decomposition-based is named similarly. In TR-RNs, each input and output channel is reshaped to 3rd order, while it is 2nd order in STR-RNs. That because the actual dimensionality in STR-Nets will be $\frac{1}{t}$ (\ie $\frac{1}{2}$) of that shown in Tab. \ref{tab:resnet dimension}.

\begin{figure}[ht]
\vskip -0.2in
\begin{center}
\subfigure[TR-RNs and STR-RNs]{
\label{fig:cr_resnet_TR} 
\includegraphics[width=0.45\columnwidth]{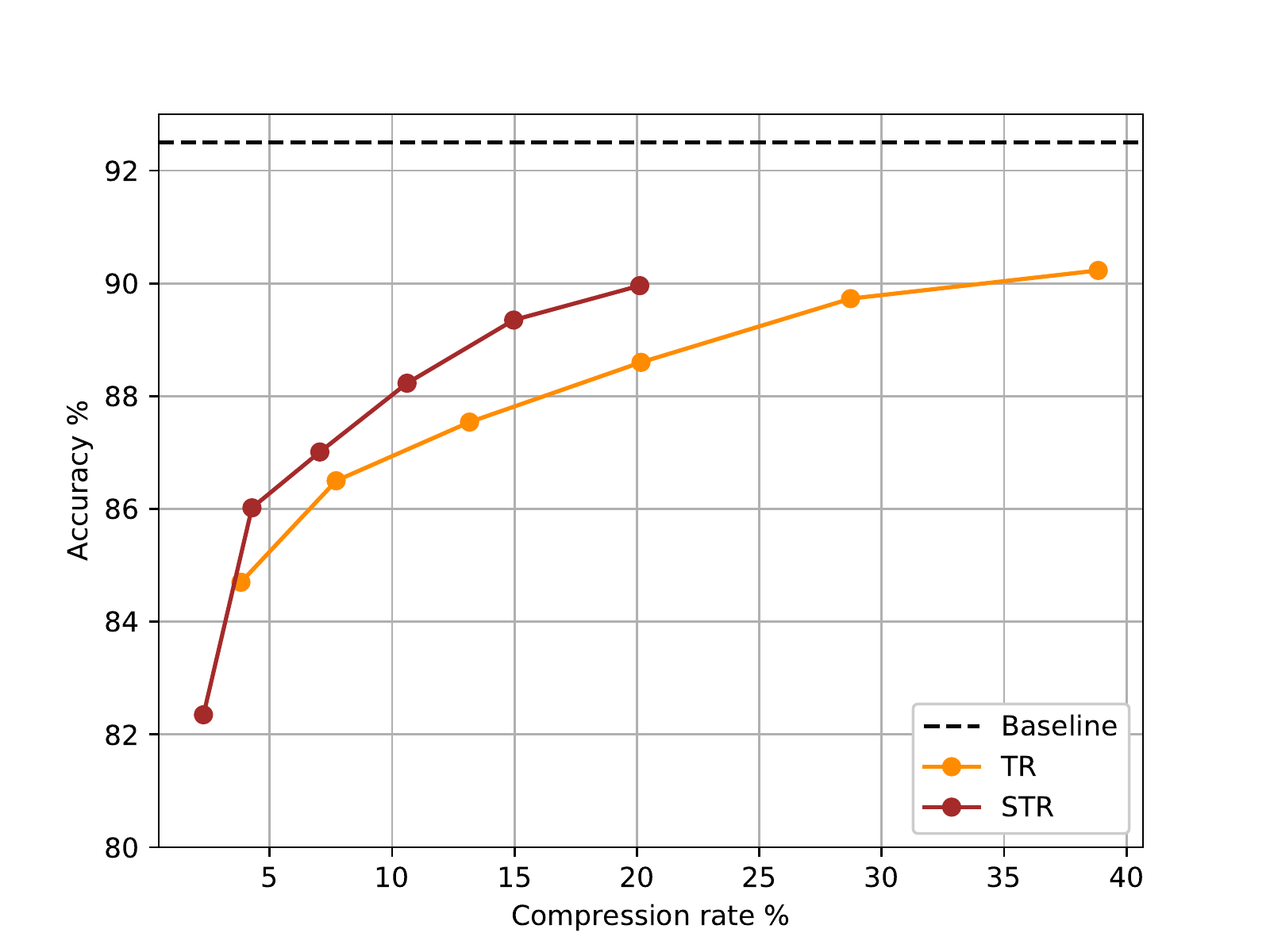}}
\subfigure[TT-RNs and STT-RNs]{
\label{fig:cr_resnet_TT} 
\includegraphics[width=0.45\columnwidth]{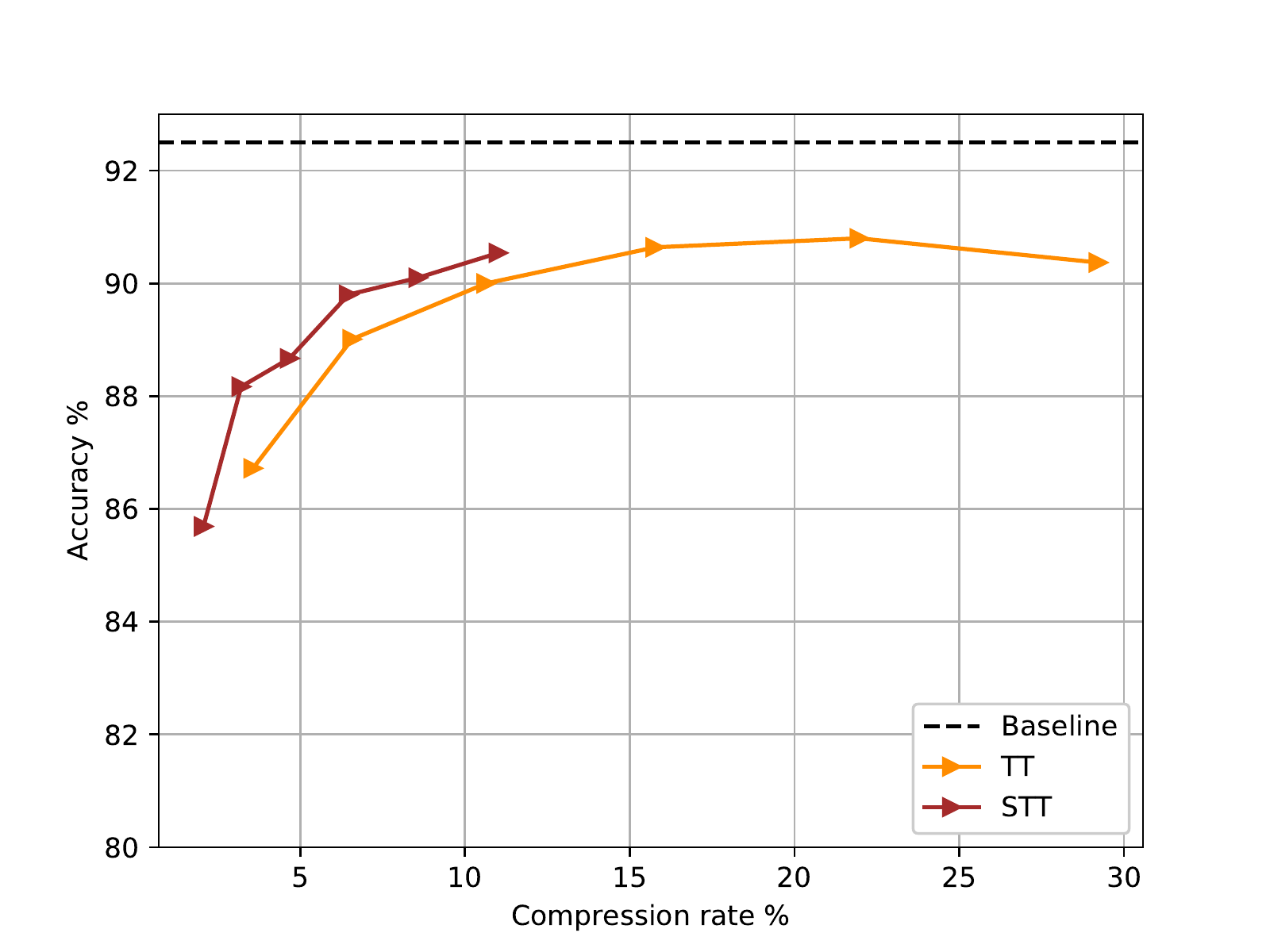}}
\caption{Accuracy vs. Model Compression Ratio for ResNet-32 on Cifar10 classification. The rank ranges from 4 to 14, with an interval of 2.Compression performance and evolution of training for ResNet-32 on Cifar10 classification.}
\label{fig:cr_resnet}
\end{center}
\vskip -0.2in
\end{figure}

\begin{table}[ht]
\caption{The deep network results. ResNet-32 on Cifar10 dataset. Training time is second per one epoch. Testing time is second per 10000 samples}
\begin{center}
\begin{tabular}{|l|c|c|c|c|c|}
\hline
 &\multicolumn{5}{c|}{Cifar10}\\
\hline
Method & Params & CF & Acc\%&Train&Test\\
\hline\hline
ResNet(RN)-32L& 0.46M & 1$\times$ &92.5&20 &2.2\\
\hline
TR-RN($R$=14) \cite{wang2018wide}& 0.18M &$2.5\times$& 90.2 & 130&12.4\\
STR-RN($R$=14) &0.09M &$5\times$ & 90.0 & 115&12.8 \\
\hline
TT-RN($R$=14) \cite{novikov2015tensorizing,garipov2016ultimate}& 0.14M &$3\times$&90.4& 149&12.6\\
STT-RN($R$=14)& \textbf{0.05M} &\textbf{9$\times$}&90.5& 126&12.6\\
\hline
Tucker-RN($R$=20) \cite{kim2015compression}& 0.16M &$3\times$&90.0& 62&4.2\\
STTu-RN($R$=20)& 0.12M &$4\times$&89.9& 39&4.2\\
\hline
\end{tabular}
\end{center}
\label{tab:resnet result}
\end{table}

Besides of STR-Nets, we design Tucker-Nets and TT-Nets based on ResNet-32 to verify that our proposed semi-tensor product-based tensor decomposition is universally effective. The reshaped data size in TT-RNs and STT-RNs is the same as that in TR-RNs and STR-RNs respectively. Fig.~\ref{fig:cr_resnet} shows that STR-RNs have higher accuracy than TR-RNs and STT-RNs have higher accuracy than TT-RNs, when their compression ratio are the same. Based on ResNet-32, TT-Nets are generally better than TR-Nets\footnote{The results of TT-Nets are different from ones in \cite{wang2018wide}, that because Wang \etal does not give a specific structure of the TT-Net, so that the TT-Net in this paper maybe different from the TT-Net in \cite{wang2018wide}.}, and semi-tensor product applied in TT decomposition can obtain better compression effect.

In Tucker and STTu decomposition-based ResNet-32, the input and output channel sizes keep original. The structure of Tucker-RNs refers to \cite{kim2015compression}. The rank size is set to 14 and 20. Table \ref{tab:resnet result} give the results of the TR-RN, the STR-RN, the TT-RN, the STT-RN, the Tucker-RN, and the STTu-RN. When the accuracy of  these nets are almost the same, the STT-RN ($R=14$)  has the highest compression factor, and the STTu-RN ($R=20$) has the fastest training speed.

\begin{figure}[ht]
\vskip -0.2in
\begin{center}
\includegraphics[width=0.5\columnwidth]{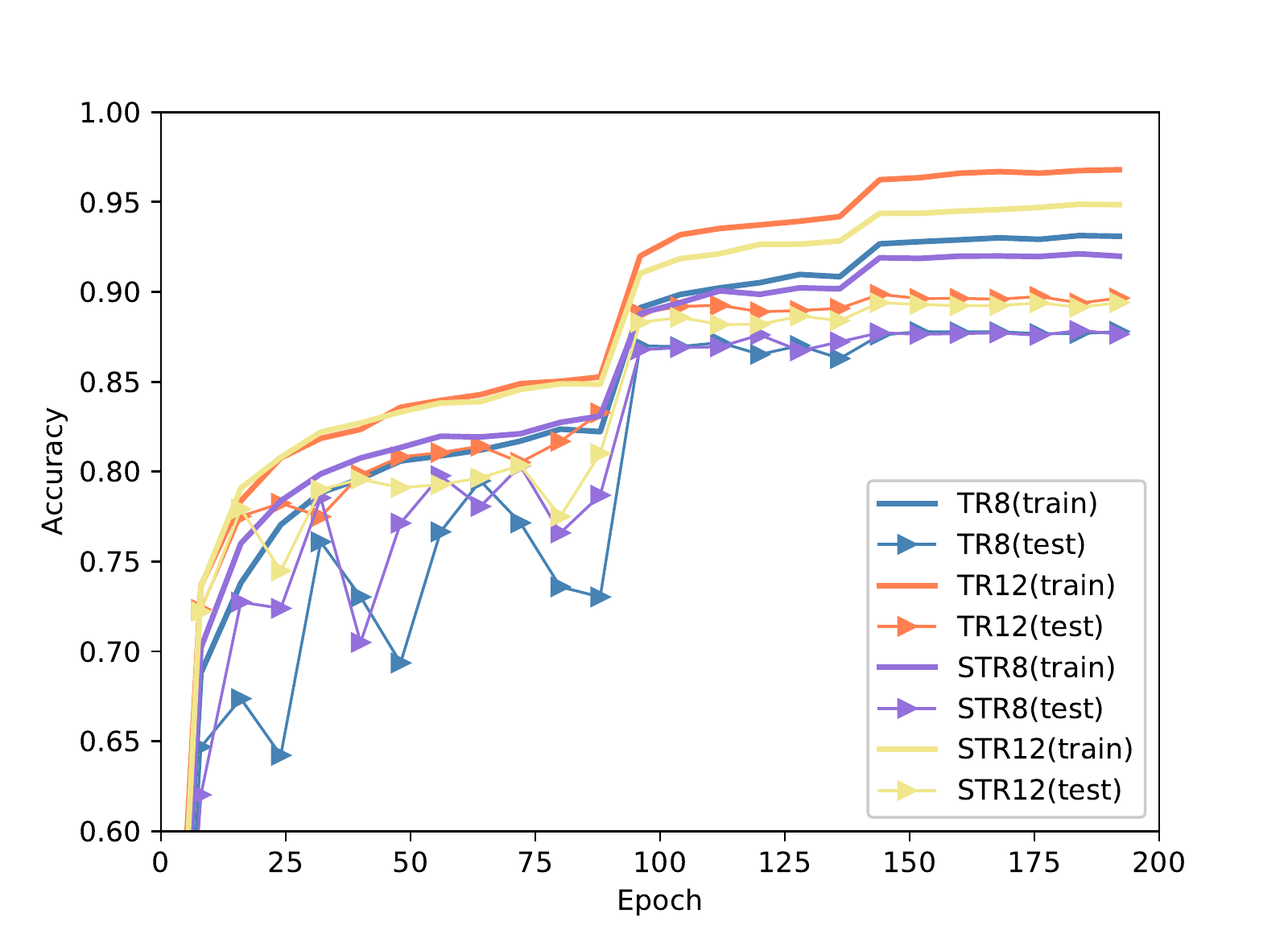}
\caption{Evolution of training compressed ResNet-32 on Cifar10, using TR-Nets and STR-Nets with different values of Rank.}
\label{fig:evolution}
\end{center}
\vskip -0.2in
\end{figure}

\textbf{Evolution.} Fig. \ref{fig:evolution} shows the training and testing errors during the training of TR-RNs and STR-RNs with $R = 8, 12$. For the same value of $R$, the generalization gap (between training and testing error) is higher for the TR-RN than the STR-RN. It suggests STR-Nets will not learn too much of the unique characteristics of the training dataset so that the generalization capability is improved.

\textbf{Training settings.} These networks are trained using stochastic gradient descent (SGD) with a multi-step learning rate policy. The batch size is set to 128 and the models trained for 200 epochs. We use the momentum of 0.9 and a weight decay of $1\times 10^{-4}$. An initial learning rate is 0.1,  which is divided by  10 after 92 and 137 epochs. The data augmentation is the same as \cite{He_2016_CVPR}.
\begin{table*}[ht]
\begin{center}
\caption{The wide network compression.}
\begin{tabular}{|l|c|c|c|c|}
\hline
 & TR-Net dimensions &Param.& STR-Nets dimensions &Param.\\
\hline\hline
ConvL1& $9\times 3\times (4\times 5\times 8)$ &$29r^2$& $9\times 3\times (10\times 16)$&$18.5R^2$\\
\hline
unit1& \tabincell{c}{$9\times (4\times 5\times 8)\times (4\times 5\times 8)$ \\$9\times (4\times 5\times 8)\times (4\times 5\times 8)$} &\tabincell{c}{$86R^2$\\$344R^2$}& \tabincell{c}{$9\times (10\times 16)\times (10\times 16)$\\$9\times (10\times 16)\times (10\times 16)$}&\tabincell{c}{$44R^2$\\$176R^2$}\\
\hline
unit2& \tabincell{c}{$9\times (4\times 5\times 8)\times (5\times 8\times 8)$ \\$9\times (5\times 8\times 8)\times (5\times 8\times 8)$} &\tabincell{c}{$94R^2$\\$408R^2$}& \tabincell{c}{$9\times (10\times 16)\times (16\times 20)$\\$9\times (16\times 20)\times (16\times 20)$}&\tabincell{c}{$49R^2$\\$216R^2$}\\
\hline
unit3&\tabincell{c}{$9\times (5\times 8\times 8)\times (8\times 8\times 10)$ \\$9\times (8\times 8\times 10)\times (8\times 8\times 10)$} &\tabincell{c}{$112R^2$\\$480R^2$}& 
\tabincell{c}{$9\times (16\times 20)\times (20\times 32)$\\$9\times (20\times 32)\times (20\times 32)$}&\tabincell{c}{$62R^2$\\$280R^2$}\\
\hline
FCL1 & $(8 \times 8\times 10) \times 10$ & $36R^2$ &$(8 \times 8\times 10) \times 10$ & $9R^2$ \\
\hline
Total & - &$1589R^2$&  - &$854.5R^2$\\
\hline
\end{tabular}
\end{center}
\label{tab:wide dimension}
\end{table*}

\begin{table*}[ht]
\begin{center}
\caption{The wide network results.}
\begin{tabular}{|l|c|c|c|c|c|}
\hline
 &\multicolumn{4}{c|}{Cifar10} & \\
\hline
Method & Params & CF & Acc\% & Train(s) & Test(s)\\
\hline\hline
WideResNet(WRN)-28L& 36.48M & 1$\times$ &95.4& 108 & 14.2 \\
\hline
TR-WRN($R$=16) \cite{wang2018wide}& 0.37M &$99\times$& 93.4 & 553& 20.6 \\
STR-WRN($R$=16) &0.20M &$\textbf{179}\times$ & 93.2 & 193& 23.8\\
\hline
TT-WRN($R$=10) \cite{novikov2015tensorizing,garipov2016ultimate}& 0.36M &$101\times$&93.4& 1304 &102.2 \\
STT-WRN($R$=10)& 0.23M &$161\times$&93.4& 873 & 89.9 \\
\hline
Tucker-WRN($R$=100) \cite{kim2015compression}& 5.10M &$7\times$&93.2& 654 &19.4 \\
STTu-WRN($R$=100)& 3.35M &$11\times$&93.4& 140 & 20.6\\
\hline
\end{tabular}
\end{center}
\vskip -0.2in
\label{tab:wideresnet result}
\end{table*}

\subsection{Deep and Wide Network}

Tensor neural networks are well-suited for wider and extremely over-parametrized models \cite{novikov2015tensorizing, wang2018wide}. Here we choose WideResNet-28, which is a deep and wide network, to verify the effectiveness of our models. The structure of WideResNet is similar to ResNet, while the output channels of WideResNet-28 (WRN) are larger than ResNet by 10. The reshaping schemes for TR-WRNs and STR-WRNs are shown in Tab. \ref{tab:wide dimension} which are the same as that in TT-WRNs and STT-WRNs, respectively. The input and output channel size in Tucker-WRNs and STTu-WRNs keep original.

WideResNet-28 classifying Cifar10 get a great performance, while the sacrifice is almost a hundred times larger memory than ResNet-32. The 3\% increase in accuracy is achieved by increasing the number of parameters from a few tenths of a million to tens of millions, and the forward and backward propagation time also increase much. As shown in Table \ref{tab:wideresnet result}, when the accuracy is almost the same, the TT-WRN and the TR-WRN have higher compression factors than the Tucker-WRN. That means the performance of TR-Nets and TT-Nets in compressing wider neural networks is better than Tucker-Nets. The TT-WRN has a compression factor similar to that of the TR-WRN when the accuracy is the same, but its training and testing time are much more. Because the computational complexity of TT-Nets is with respect to $IO$ while that of TR-Nets is with respect to $(I+O)$, as shown in Table~\ref{tab:STR-ConvL} and Table \ref{tab:STT-ConvL}. $I$ and $O$ are the numbers of input channels and output channels, respectively, both of them represent the width of convolutional neural networks. Thus TR decomposition is more suitable for compressing wide neural networks than TT and Tucker decomposition.

We apply STR, STT and STTu in WideResNet-28, and all of them achieve good performance. We choose the rank size $R = 16, 10, 100$ for the TR-WRN, the TT-WRN and the Tucker-WRN respectively, and their corresponding semi-tensor based counterparts share the same rank size. In this way, The above six networks obtain the similar accuracy which is 2\% less than the original network. The compression factor of the STR-WRN is $80 \times$ more than the TR-WRN; the STT-WRN is $60 \times$ more than the TT-WRN; the STTu-WRN is $4 \times$ more than the Tucker-WRN. The training times of all the STP-based tensor neural networks are less than that of their corresponding tensor neural networks. Especially for the TR-WRN and the Tucker-WRN,   the problem of long training time caused by the increase of network width is effectively alleviated.


\section{Conclusion}
In this paper, we generalize tensor products by replacing matrix products with semi-tensor products in tensor networks, which results in a group of new tensor decompositions. The obtained decompositions are mainly used for deep neural networks compression. Our experiments have demonstrated that the proposed STP-Nets (\ie STR-Nets, STT-Nets, STTu-Nets) achieve a  
superior compression ratio and training speeding with the same classification accuracy, compared with corresponding conventional tensor neural networks. 
The STP operation removes the dimension consistency limitation of tensor modulus product, and expresses the same information in a more compact structure with less memory. However, our method requires the proportional relationship between factors, we will generalize it to the third case (arbitrary sizes) in future work.

\ifCLASSOPTIONcompsoc


\ifCLASSOPTIONcaptionsoff
  \newpage
\fi



%


{\small
\bibliographystyle{ieee_fullname}
\bibliography{egbib}
}
%








\end{document}